\documentclass{article}

\usepackage{arxiv}

\usepackage[utf8]{inputenc} % allow utf-8 input
\usepackage[T1]{fontenc}    % use 8-bit T1 fonts
\usepackage{hyperref}       % hyperlinks
\usepackage{url}            % simple URL typesetting
\usepackage{booktabs}       % professional-quality tables
\usepackage{amsfonts}       % blackboard math symbols
\usepackage{nicefrac}       % compact symbols for 1/2, etc.
\usepackage{microtype}      % microtypography
\usepackage{lipsum}
\usepackage{fancyhdr}       % header
\usepackage{graphicx}       % graphics
\graphicspath{{media/}}     % organize your
\usepackage{microtype}
\usepackage{amsmath}
\usepackage{amssymb}
\usepackage{graphicx}
\usepackage{subfigure}
\usepackage{enumitem}
\usepackage{booktabs} % for professional tables
\usepackage[dvipsnames]{xcolor}
\usepackage{float}
\usepackage{adjustbox}
\usepackage{xspace}
\usepackage{setspace}
\usepackage{natbib}

\usepackage{hyperref}
\newcommand\myshade{50}
\colorlet{mylinkcolor}{blue}
\colorlet{mycitecolor}{green}
\colorlet{myurlcolor}{red}
\hypersetup{
  linkcolor  = mylinkcolor!\myshade!black,
  citecolor  = mycitecolor!\myshade!black,
  urlcolor   = myurlcolor!\myshade!black,
  colorlinks = true,
}

\newcommand{\RR}{\mathbb{R}}

\newcommand{\Oo}{\mathcal{O}}

\renewcommand{\phi}{\varphi}

\DeclareMathOperator{\softmax}{softmax}
\DeclareMathOperator{\topk}{topk}

\newcommand{\foralls}{\forall \,}
\renewcommand{\leq}{\leqslant}
\renewcommand{\geq}{\geqslant}

\renewcommand{\epsilon}{\varepsilon}
\renewcommand{\imath}{\mathrm{i}}

\newcommand{\dotp}[2]{\langle #1,\,#2\rangle}

\newcommand{\abs}[1]{\left\lvert#1\right\rvert} % modified by Vincent

\DeclareMathOperator*{\argmax}{argmax}
\newlength{\restsubwidth}
\newlength{\restsubheight}
\newlength{\restsubmoreheight}
\newcommand{\rest}[2]{%
        \settowidth{\restsubwidth}{\ensuremath{#2}}
        \settoheight{\restsubheight}{\ensuremath{{}_{#2}}}
        \ensuremath{{#1\hskip 0.5pt}_{\vrule\kern2pt\parbox[b][%
        4pt][b]{\the\restsubwidth}{%
                        \ensuremath{{}_{#2}}}}}
        }

\newcommand{\base}{\textsc{s-base}\xspace}
\newcommand{\baseorig}{\textsc{base}\xspace}
\newcommand{\rlr}{\mbox{\textsc{rl-r}}\xspace}
\newcommand{\hash}{\textsc{hash}\xspace}
\newcommand{\smoe}{\textsc{smoe}\xspace}
\newcommand{\smoes}{\textsc{smoe}s\xspace}
\newcommand{\epc}{\textsc{epc}\xspace}

\newcommand{\Emax}{E_{\max}}
\DeclareMathOperator{\start}{start}
\newcommand{\Estart}{E_{\start}}
\newcommand{\Ncut}{N_{\textrm{cutoff}}}

\newcommand{\wikitext}{\textit{WikiText-103}\xspace}
\newcommand{\cc}{\textit{Curation Corpus}\xspace}
\newcommand{\lambada}{\textit{LAMBADA}\xspace}
\newcommand{\pile}{\textit{The Pile}\xspace}
\newcommand{\cfour}{\textit{C4}\xspace}

\makeatletter
\renewcommand{\sectionautorefname}{\S\@gobble}%
\renewcommand{\subsectionautorefname}{\S\@gobble}%
\renewcommand{\subsubsectionautorefname}{\S\@gobble}%
\makeatother
\usepackage{changepage}                 % adjust margins for selected portions
\newlength{\offsetpage}
{\end{adjustwidth}}

\title{Unified Scaling Laws for Routed Language Models
}
\date{\vspace{-5ex}}
\usepackage{ragged2e}

\author{
  {\setstretch{1.2}

  \textbf{Aidan Clark$^*$, Diego de las Casas$^*$, Aurelia Guy$^*$, Arthur Mensch$^*$}
  
  \vspace{.5em}
  
    \justifying
  \textbf{Michela Paganini, Jordan Hoffmann, Bogdan Damoc, Blake Hechtman$^\ddagger$, Trevor Cai, Sebastian Borgeaud,}
  \linebreak
  \textbf{George van den Driessche, Eliza Rutherford, Tom Hennigan, Matthew Johnson$^\ddagger$, Katie Millican,}
  \linebreak
  \textbf{Albin Cassirer, Chris Jones, Elena Buchatskaya, David Budden, Laurent Sifre, Simon Osindero,}
  \linebreak
  \hspace*{7em} \textbf{Oriol Vinyals, Jack Rae, Erich Elsen, Koray Kavukcuoglu, Karen Simonyan}
  
  \vspace{.5em}
  \centering
  \hspace*{2.4em}DeepMind \quad\quad\quad\quad\quad Google Research$^\ddagger$
  }
}

\begin{document}
\maketitle

\def\thefootnote{}\footnotetext{Correspondence to \text{aidan.b.clark@gmail.com}, \text{diegolascasas@deepmind.com}. All affiliation to DeepMind unless noted.}\def\thefootnote{\arabic{footnote}}
\def\thefootnote{*}\footnotetext{Shared first authorship.}\def\thefootnote{\arabic{footnote}}
\begin{abstract}
The performance of a language model has been shown to be effectively modeled as a power-law in its parameter count.
Here we study the scaling behaviors of \textit{Routing Networks}: architectures that conditionally use only a subset of their parameters while processing an input.
For these models, parameter count and computational requirement  form two independent axes along which an increase leads to better performance.
In this work we derive and justify scaling laws defined on these two variables which generalize those known for standard language models and describe the performance of a wide range of routing architectures trained via three different techniques.
Afterwards we provide two applications of these laws: first deriving an \textit{Effective Parameter Count} along which all models scale at the same rate, 
and then using the scaling coefficients to give a quantitative comparison of the three routing techniques considered.
Our analysis derives from an extensive evaluation of Routing Networks across five orders of magnitude of size, including models with hundreds of experts and hundreds of billions of parameters.
\end{abstract}

\section{Introduction}
\label{introduction}

It is a commonly held belief that increasing the size of a neural network leads to better performance, especially when training on large and diverse real-world datasets.
This vague and debated notion has become increasingly justified as large empirical studies have shown that the performance of models on many interesting classes of problems are well understood as power-laws; where a multiplicative increase in model size leads to an additive reduction in the model's loss \citep{kaplan2020scaling, hernandez2021scaling, henighan2020scaling, rosenfeld2019constructive}. These relationships are not well understood, but a key implication is that a sequence of small\footnote{Measured as training or inference floating point operations, devices or time required, financial cost, carbon emissions, etc.} models can be used both to infer the performance of models many times more powerful, but also to provide global information about the scalability of an architecture.

Enter Routing Networks: models with the unusual property that each input interacts with only a subset of the network's parameters --- chosen independently for each datapoint \citep{bengio2015conditional, bengio2013estimating, denoyer2014deep}. For a Routing Network, the number of parameters is nearly independent from the computational cost of processing a datapoint. This bifurcates the definition of size and prevents a scaling law in parameters alone from fully describing the model class.
Specific Routing Networks have been trained successfully at large scales \citep{fedus2021switch, du2021glam, artetxe2021efficient}, but the general scaling behavior is not well understood. In this work we analyze the behavior of routed language models so that we might infer the scaling laws that describe their performance.

\pagebreak

\begin{figure*}[t]
  \centering
  \includegraphics[width=\textwidth]{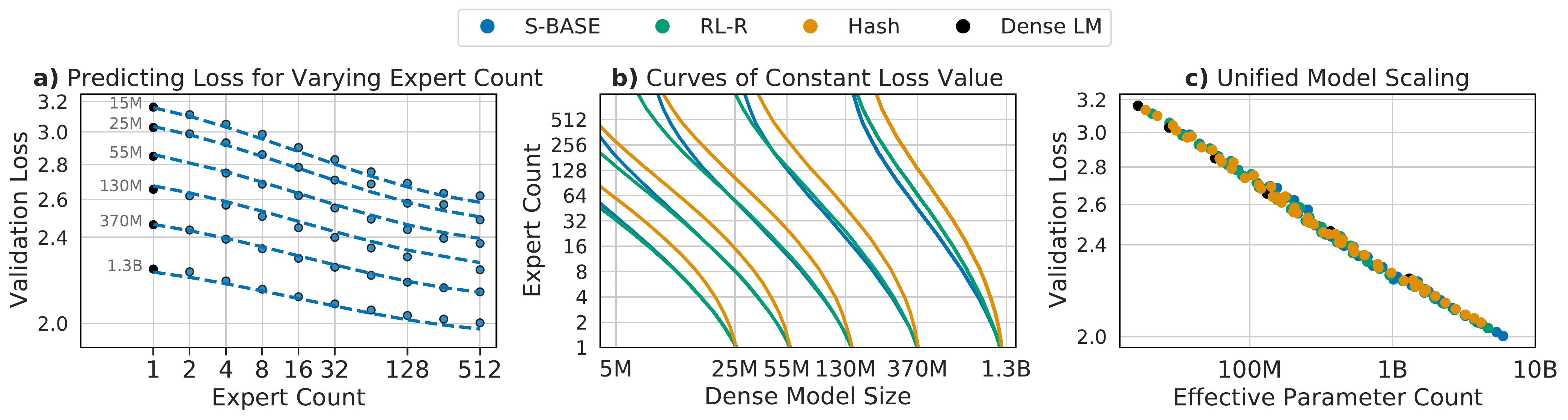}
  \vspace{-5mm}
\caption{
% \textbf{Scaling law in parameters and compute}:
\textbf{(a)} The performance achieved by Routing Networks when varying the number of experts for a fixed dense model size is described by a bilinear function (Eq.~\ref{eq:real_joint_scaling_law}), \textbf{(b)} whose level curves indicate how to trade model size with expert count to maintain a fixed performance, \textbf{(c)} and which can be manipulated to align dense and routed model performance under a shared power law.}\label{fig:main}
\end{figure*}

\paragraph{Key contributions.} We analyze three different techniques for training Routing Networks, detailed in \autoref{sec:ref}: \mbox{Sinkhorn-\baseorig}, a sparse mixture-of-experts (\smoe) approach modifying \baseorig \citep{lewis2021base}; non-parametric \hash Layers \citep{roller2021hash}; and routing via Reinforcement Learning (\rlr). With models up to 200 billion parameters, we observe the following:

\begin{enumerate}[wide,itemsep=0pt,topsep=0pt, labelindent=3pt]
    \item Routing improves the performance of language models across all sizes and variants attempted (see \autoref{fig:main}).
    \item Training a Routing Network with RL (\autoref{subsec:rl}), a technique used in early routing work \citep{bengio2013estimating}, is of comparable effectiveness to state-of-the-art techniques.
    \item The performance of all Routing Networks is accurately described by scaling laws in the number of experts and in the underlying dense model size (\autoref{sec:scaling}) which generalize those from \citet{kaplan2020scaling}.
    \item These laws can be restated in terms of parameter count and inference compute, capturing an even wider set of routing architectures under a shared fit (\autoref{sec:generalizations}).
    \item They further imply an \textit{Effective Parameter Count}: a mapping equating the performance and scaling for both dense and routed networks (\autoref{sec:applications}).
\end{enumerate}

\section{Background}

We first review the language modelling problem and existing scaling laws before discussing the process of routing a neural network and how it is applied to language models.

\paragraph{Language modelling.}
We consider the problem of autoregressively predicting natural language, a task with consistent and predictable scaling characteristics across many orders of magnitude \citep{henighan2020scaling, kaplan2020scaling}.
The objective is to maximize the likelihood of a sequence of tokens $P(x_1, \dotsc , x_T)$ factored auto-regressively as $p \left (x_1, \dotsc , x_T \right ) = \prod_i^T p \left (x_i | x_{j < i} \right )$.
Our primary metric of performance is the negative log-likelihood of a validation dataset whose statistics match the training distribution. We focus on this validation loss, but briefly consider zero-shot transfer to other tasks in \autoref{apsec:transfer}.

\paragraph{Scaling laws for large-scale data.} We train on a multi-trillion-token compendium of English language text comprising documents from the internet alongside open-source text datasets, details of which are given in \citet{rae2021scaling}. In this setting \citet{kaplan2020scaling} argue that the converged performance of a model trained on a dataset of infinite size is a power-law in the model's parameter count~$N$. Our dataset is not infinite, but its size -- and the lack of any observed overfitting -- make this a reasonable approximation. We consider the final (and best) evaluation value as the converged value, though this is also an approximation which is discussed further in~ \autoref{apsec:convergence}.

\subsection{Routing Networks}
\label{sec:routing}

 Power-law scaling implies the performance of a language model increases with size, but so too does the compute needed to train the model. This undesirable connection between size and computation motivates a search for architectures wherein the two are disentangled. Routing Networks are one such class of model: a type of neural
network that incorporates a specific flavor of conditional
computation. In a Routing Network, each input (e.g., a token of text) is transformed into an output while only interacting with a fixed subset of the network's parameters -- dynamically selected based on the input itself. Many sparsely-activated networks have this property, but here we exclusively study the layout based on  Sparse Mixtures of Experts \citep{shazeer2017outrageously} where multiple sub-components of a deep neural network (i.e., several layers) are independently converted to routed equivalents and jointly trained with the rest of the network.

\paragraph{Routing a single layer.} The core idea of a routed layer is that multiple versions of the parameters are kept, and a per-input decision on which version to use is made. To route a layer $f_\theta$ in $E$ ways, we start by creating $E$ separate versions of the parameters $\theta$ ($\{\theta_1, ... \theta_E\}$) where $f$ using the $i$-th version of the parameters ($f_i\triangleq f_{\theta_i}$) is termed the \textit{i-th Expert}. To determine which expert to pick given the input, we introduce an additional \textit{router} function  $\rho: \mathbb{R}^M \to \left[1,E\right]$ associated to the layer, typically a small network itself, with parameters $\phi$. The routed form $h$ of $f$ is then given by $h(x) \triangleq f_{\rho(x)}(x)$.
When performance increases with $E$, routing gives a method by which to improve a neural network with minimal computational increase (corresponding only to the compute needed by $\rho(x)$).

We also consider the $K$-way routed generalization, where the router outputs a set of integers as $\rho(\cdot): \mathbb{R}^M \to \left[1,E\right]^K$, and we set the output of the layer to be the sum of the outputs of each expert, namely $h(x) \triangleq \sum_{i \in \rho(x)} f_i(x)$. We default to $K=1$, but revisit this in \autoref{sec:generalizations}.

\paragraph{Routed Transformers}
We apply routing to a decoder-only Transformer \citep{vaswani2017attention} to measure the scaling properties that result: an architecture chosen due to its state-of-the-art performance. Details of the baseline architecture we use are in \autoref{app:archtecture}. We will refer to non-routed Transformers as \textit{dense} models, in opposition to Routed Transformers which \textit{sparsely} activate some of their parameters. Our conversion to a Routed Transformer is the same as is used in prior work \citep{lepikhin2020gshard, fedus2021switch}. Namely, we apply routing to every other set of feedforward components (FFWs) of the Transformer, sub-components that act on each timestep independently. Though different layers can have
different numbers of experts, here all routed
layers share the same number of experts $E$, and we will refer
to the network as being \textit{routed $E$ ways}.

\paragraph{Model size and inference cost.}
We use $N$ to indicate a network's \textit{dense model size}: the number of parameters any one input interacts with. This is in opposition to $P$: the total number of parameters. For a dense model, $P = N$, whereas for a Routing Network $P$ is roughly proportional to $N\cdot E$, with factors that depend on details of the routing architecture (\autoref{sec:generalizations}).
Except for a small overhead due to running the routers, the cost $F$ (in TeraFLOPs) of executing a Routed Transformer is the same as its dense equivalent.

\paragraph{Training Details.}\label{ssec:training_details}
All models are trained on TPUs with JAX \citep{bradbury2020jax} using a combination of data, expert (see \autoref{apsec:routing}) and sharding parallelism \citep{shoeybi2019megatron}. Models were trained with a sequence length of $2048$ and batch size of $256$ for 250,000 steps, i.e. 130 billion tokens, regardless of $N$. This is an important detail, and we discuss some of the implications in \autoref{apsec:convergence}. All were optimized with AdamW \citep{loshchilov2018decoupled} and ZeRO Stage 1 was used to shard the optimizer state \citep{rajbhandari2020zero}. Appendix \ref{app:archtecture} contains further details.

\section{Routing Techniques}
\label{sec:ref}

If the benefit of Routing Networks is the decoupling of parameter capacity from network cost, the fundamental difficulty is in effectively learning the parameters $\phi$ of the router given the non-differentiability of its output. Much research in Routing Networks has therefore focused on techniques for learning $\phi$. A major finding of this work is that three notably different techniques of training Routing Networks are effectively described by the same scaling laws. We now introduce and contextualize these three methods.

\subsection{Sparse Mixture-of-Experts via Weighting}

Sparse Mixture-of-Experts (\smoe) methods \citep{shazeer2017outrageously} solve the problem of non-differentiability by reusing the probability of expert selection as a scalar multiplier on that expert's output, guaranteeing a gradient passed to the logits of selected experts despite the the non-differentiability of sampling from those logits. Formally, the router is given as $\rho(x) = \topk(Wx + b)$, where $Wx + b$ is an unnormalized distribution over $\left[1,E\right]$ from which the experts corresponding to the top $K$ values are selected. In the final output of the routed layer, the normalized logits are reused as \textit{gating weights}, i.e. the final output of the routed layer is $h(x) = \sum_{i \in \rho(x)} g_i(x) f_i(x)$ where $g(x) = \softmax(Wx + b)$.

Though this formulation supplies a gradient to $\phi = (W,b)$, it represents changes to the scalar multiplier and does not directly correspond to optimizing expert selection. This method is nevertheless effective, and can be seen as a sparse approximation to dense mixture of experts models \citep{eigen2013learning, jacobs1991adaptive} where the likelihood of skipping an expert is inversely proportional to the value of its scalar gate $g_i$.

It was conjectured that \smoes require $(K \geq 2)$-way routing to produce effective gradients in the routers \citep{shazeer2017outrageously}, and many attempts at incorporating routing into large Transformers use $K = 2$ \citep{lepikhin2020gshard, du2021glam}. However recently this has been challenged, and  stable modifications have been proposed for $K = 1$; namely the Switch Transformer \citep{fedus2021switch}. Most \smoes, including Switch, are reliant on auxiliary balancing losses which encourage the router output $\rho(x)$ to be more uniform across minibatches of inputs. To improve on this, \baseorig \citep{lewis2021base} post-processes the router output with a Hungarian Matching algorithm that re-assigns expert selections to ensure that all experts are selected evenly.

Our implementation of \baseorig replaces the Hungarian Matching with a regularized Optimal Transport formulation \citep{cuturi2013sinkhorn} using the Sinkhorn algorithm as an approximate matching step during expert selection. This substantially improves routing efficiency on accelerated hardware (details in \autoref{sec:sinkhorn}). We call the resulting method Sinkhorn-\baseorig (\base), and use it as the representative of \smoe methods, as early tests showed the benefit of its balancing mechanism. 

\subsection{Input-based Deterministic Hash Routing}

\begin{figure*}[t!]
  \centering \includegraphics[width=\linewidth]{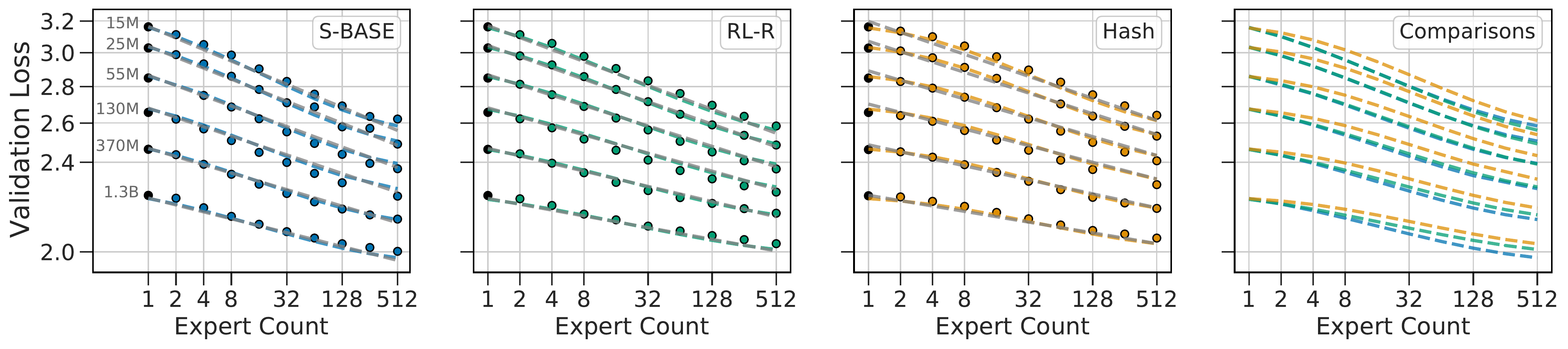}
\caption{Validation losses with fits from Equation \ref{eq:real_joint_scaling_law} plotted as a dotted line for \base, \hash and \rlr respectively. On the right, the prediction curves for all model sizes and all techniques overlapping to show relative performance. Fits to Eq.~\eqref{eq:no_tailoff} are overlaid in grey.}
\label{fig:joint_curves}
\end{figure*}

An alternative approach eschews extra parameters completely and represents $\rho$ as a fixed function of the input. This is the concept pioneered by \hash Layers \citep{roller2021hash} which circumvents the need to simultaneously learn $\phi$ and $\theta$. Our implementation takes the token ID assigned to the input by the SentencePiece tokenizer \citep{kudo2018sentencepiece} and uses the remainder of it divided by $E$ as the expert selection. See~\autoref{sec:hash_details} for details.

\subsection{Routing via Reinforcement Learning}\label{subsec:rl}

Finally, we re-analyze a technique that optimizes the router via Reinforcement Learning 
(a class of methods we call \rlr), which was proposed in early work on neural conditional computation \citep{bengio2013estimating, bengio2015conditional, bengio2017reinforcement, denoyer2014deep}. In this approach each router is seen as a policy whose actions are the selection of an expert in each routed layer and whose observations are the activations passed to that router. After completing the forward pass, the probability the Routed Transformer assigns to the correct output token can be used as a reward, maximization of which is equivalent to minimization of NLL. To jointly train the experts and the router, we minimize a composite loss formed with the language modelling loss and a policy-gradient term \citep{sutton2000policy} using the selected set of experts as actions. We highlight that the optimal expert selection is dependent not only on the input activations but on the parameters of the rest of the network. This disrupts the theoretical underpinning, crucial to RL, that this is a Markov Decision Process. Nevertheless, it has been observed that this theoretical issue does not affect the practicality of the method \citep{rosenbaum2019routing}.

Relative to \smoe, \rlr benefits from directly optimizing actions to improve the language modelling loss. However this absence of bias comes with complications, especially the high variance of the gradient \citep{rosenbaum2019routing, denoyer2014deep}. We use \textsc{reinforce} with a learned baseline \citep{williams1992simple, sutton2018reinforcement} to address this issue, so that improving the policy means increasing the likelihood of selecting experts which lead to a better than average next token prediction. As with \smoe, we find it useful to add a balancing term. To our knowledge, we are the first to experiment routing with Reinforcement Learning on large Transformer-based language models---we therefore explore key ablations in Appendix \ref{apsec:rlr_variants}.

\section{Scaling Behavior at Convergence}
\label{sec:scaling}

Our main hypothesis is that the converged log-loss of a Routing Network is bilinear in the terms $\log N$ and $\log \widehat{E}$, where $\hat E$ is a saturating transformation of $E$. Specifically, we fit the 6-parameter scaling law:
\begin{align}
       &\!\!\!\log L (N, E) \triangleq a \log N {+} b \log \widehat{E} {+} c \log N \log \widehat{E} {+} d \label{eq:real_joint_scaling_law}\\
&\text{where}\quad    \frac{1}{\widehat{E}} \triangleq
        \frac{1}{E - 1 + \left(\frac{1}{E_\text{start}} - \frac{1}{E_{\max}}\right)^{-1}} + \frac{1}{E_{\max}}.\notag
\end{align}
We can generalize this law across a wider range of routing architectures by a change of variables, using the model inference cost $F$ and the total number of parameters~$P$, as:
\begin{align}
       \log L (F, B) \triangleq a \log F {+} b \log \widehat{B} {+} c \log F \log \widehat{B} {+} d,\label{eq:real_joint_scaling_law_fp}
\end{align}

\begin{table}[t]
\centering
\begin{minipage}{.6\linewidth}
\caption{Leave-One-Out RMSLE Fit in $(N, E)$. The last row is \linebreak computed for each model size independently; this gives an lower \linebreak bound of the error of any joint scaling law.}\label{tab:rmse}
\begin{tabular}{ c c | c c c}
 \toprule
 $L$ log-log prediction & Eq. & \base & \rlr & \hash \\
 \midrule
 Separably linear in $N$, $E$ & \eqref{eq:separable} & 80e-4 & 90e-4 & 90e-4 \\
 Bilinear in $(N, E)$ & \eqref{eq:no_tailoff} & 60e-4 & 57e-4 & 60e-4 \\
 Bilin. + saturat. in $(N, E)$ & \eqref{eq:real_joint_scaling_law} & 58e-4 & 56e-4 & 56e-4 \\
 \midrule
 Per-$N$ fits in $(E)$ & \eqref{eq:hopeful_power} & \textbf{46e-4} & \textbf{29e-4} & \textbf{19e-4} \\
  \bottomrule
\end{tabular}

\end{minipage}%
\begin{minipage}{.4\linewidth}
\centering
\caption{Dense scaling values (see also~\autoref{apsec:convergence}).}\label{tab:joint_alphas}
        
\begin{tabular}{ c | c c}
%  \toprule
 & $\alpha_N$ & $N_c$ \\
 \midrule
  \textbf{Ours} & $0.078$ & $3.568e13$\\
%   \midrule
  \textbf{\citet{kaplan2020scaling}} & $0.076$ & $8.8e13$ \\
  \bottomrule
\end{tabular}

    \end{minipage} 
\end{table}

where $B \triangleq \frac{P}{F}$ and $B \to \hat B$ is the same saturating transform as $E \to \hat E$. Before justifying Equation~\eqref{eq:real_joint_scaling_law}, we validate its candidacy by fitting it to empirical data obtained on a large sweep of models. This sweep consists of a Routing Network trained for each of the three techniques described in \autoref{sec:ref}: across six model sizes (described in \autoref{tab:architectures}) while varying $E$ across $[2, 4, 8, 16, 32, 64, 128, 256, 512]$.
This totals 168 different models, including dense baselines.

The observed losses for each model are shown in \autoref{fig:joint_curves}(a-c). We fit Eq.~\eqref{eq:real_joint_scaling_law} to each routing method and plot predictions for fixed values of $N$ as dotted lines. The goodness-of-fit across all methods is apparent, as is the clear behavior that increasing $E$ leads to a reduction in validation loss. \autoref{fig:joint_curves}(d) plots the relative predictions for all three techniques, clearly showing that \base performs best across all model sizes, followed by \rlr, followed by \hash (see~\autoref{sec:comparisons}).
The remainder of this section justifies the chosen functional forms~\eqref{eq:real_joint_scaling_law} and~\eqref{eq:real_joint_scaling_law_fp}; first supposing independent power laws in $N$ and $E$ (\autoref{section:conditional_scaling}), then introducing a multiplicative interaction (\autoref{subsec:quadratic}) and saturation in the second term (\autoref{sec:tail_off}), followed by a change of variables (\autoref{sec:generalizations}). The benefit gained by this progression of fits can be seen in~\autoref{tab:rmse}. Notations are recalled in \autoref{fig:alpha_es}.

\subsection{Separable Scaling Laws in Model Size and Experts}
\label{section:conditional_scaling}

% \paragraph{Scaling in $N$.}
\citet{kaplan2020scaling} argue that the converged performance of a dense model with $N$ parameters can be modelled accurately as the two-parameter power law
\begin{equation}\label{eq:dense_law}
    \log L(N) \triangleq a \log N + d,
      \quad\text{i.e.}\quad 
       L(N) = {\left(\frac{N_c}{N}\right)}^{\alpha_N}
\end{equation}
where $\alpha_N \triangleq -a$ and $N_c \triangleq 10^{d/-a}$. We can re-estimate these coefficients from the performance of our own dense models, leading to estimations in \autoref{tab:joint_alphas}. The similarity of $\alpha_N$ is a reassuring sanity check (there are differences in dataset, vocabulary, tokenization and model which effect $N_c$).

\begin{figure}[t]
\centering
\hfill
\begin{minipage}{0.3\linewidth}
  \includegraphics[width=\linewidth]{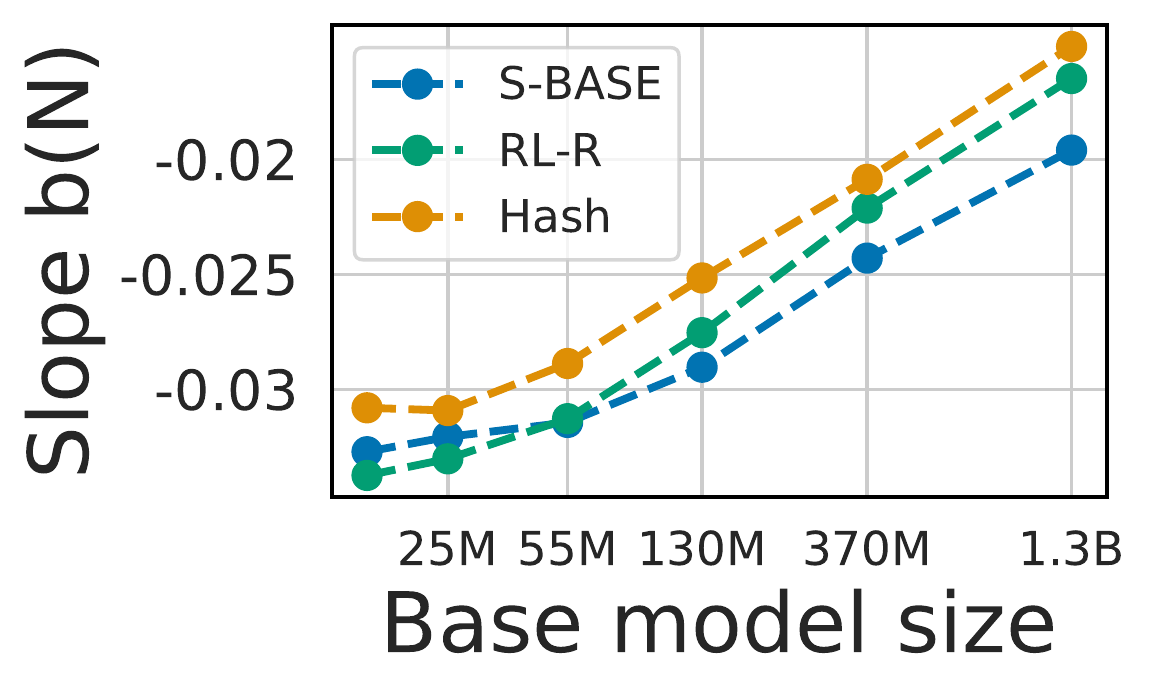}
\end{minipage}
\hspace{2em}
\hfill
\begin{minipage}{.45\linewidth}
% \small
  \begin{tabular}{l|l}
%   \toprule
  $N$ & Parameter Count in Base Model \\
  $E$ & Number of Experts \\
%   \midrule
  $P$ & Total Number of Parameters  \\
  $F$ & Compute per Inference (in TeraFLOPs) \\
  $B$ & Parameter Utilization Ratio \\
%   \midrule
  $\bar N$ & \epc: The Effective Parameter Count \\
  \end{tabular}
    \vfill
\end{minipage}
\caption{\textit{Left}: $b(N)$ increases with $N$. \textit{Right}: Notations.}\label{fig:alpha_es}
\end{figure}

An immediate hypothesis is that for all values of $N$, scaling in $E$ obeys a similar power law:
\begin{equation}
    \log L_N(E) \triangleq b \log E + d'
    \label{eq:hopeful_power}
\end{equation}
Because $L_N(1) = L(N)$ (a fact we will call \textit{dense equivalence}),
\eqref{eq:dense_law} and~\eqref{eq:hopeful_power} can be combined into:
\begin{equation}
\log L_N(E) \triangleq a \log N + b \log E + d,\label{eq:separable}
\end{equation}
corresponding to the multiplicative separated power law:
\begin{equation}
    L_N(E) = \left( \frac{10^{d/a}}{N}\right)^{a} \left( \frac{1}{E} \right)^{b}\label{eq:multi_separable}
\end{equation}
If Eq.~\eqref{eq:hopeful_power} fits observed data for any $N$ we can proceed with an assumption that scaling in $E$ obeys a power-law for fixed $N$. Observing a constant $b$ across $N$ would allow to fit Eq.~\eqref{eq:separable} to models ranging across $N$ and $E$ simultaneously.
\paragraph{Fitting.}
The first hypothesis is easily tested and confirmed to a reasonable degree. We fit Eq.~\eqref{eq:hopeful_power} for each technique and value of $N$ separately, plotted as colored lines in \autoref{fig:linear_fits}. The values of $b$ are shown in \autoref{fig:alpha_es}.

We observe that $b(N)$ is \textit{increasing} with $N$ (values listed in \autoref{tab:alphas}), corresponding to a reduction in benefit from routing as size increases, with a slope that is approximately linear in $\log N$ (\autoref{fig:alpha_es}). Eq.~\eqref{eq:separable} requires that $b$ remains fixed across $N$; therefore we expect it to poorly predict model performance. We can attempt a fit nevertheless: plotted in grey in \autoref{fig:linear_fits}. Qualitatively, this mis-predicts some validation losses by over 0.2, particularly overestimating the performance at large $N$ and $E$. As reported in~\autoref{tab:rmse}, the fit has held-out RMSLE values greater than 80e-4.

\subsection{Quadratic Interaction in $N$ and $E$}
\label{subsec:quadratic}
\begin{figure}[t]
    \centering
  \includegraphics[width=0.3\linewidth]{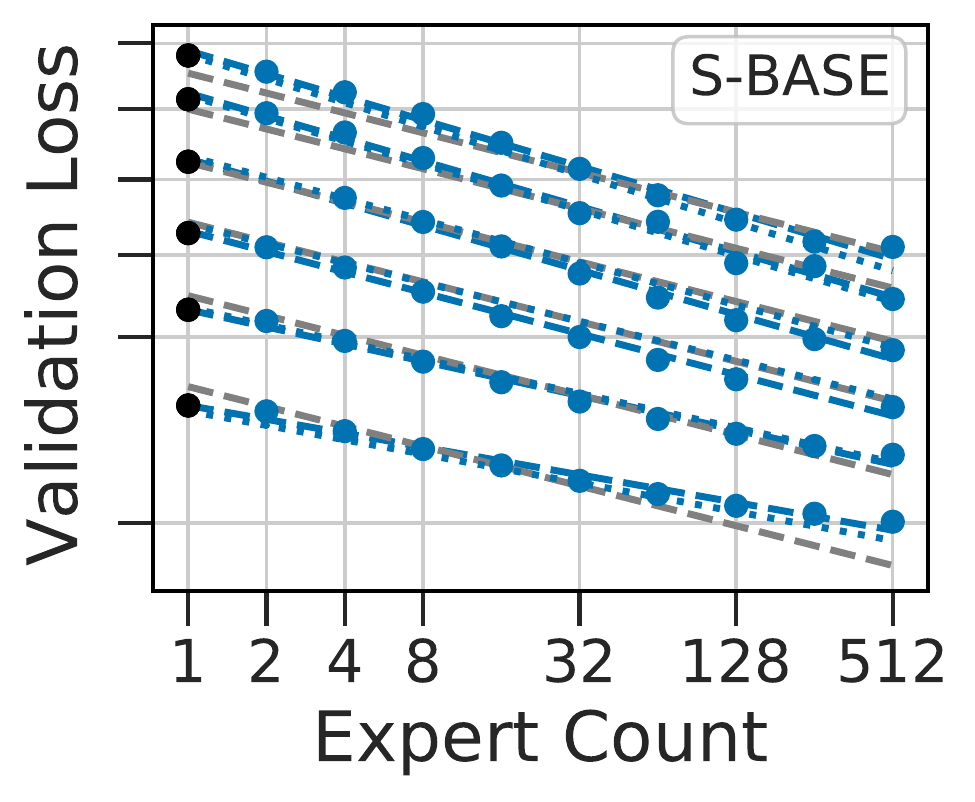}%
  \hspace{-0.3em}%
  \includegraphics[width=0.3\linewidth]{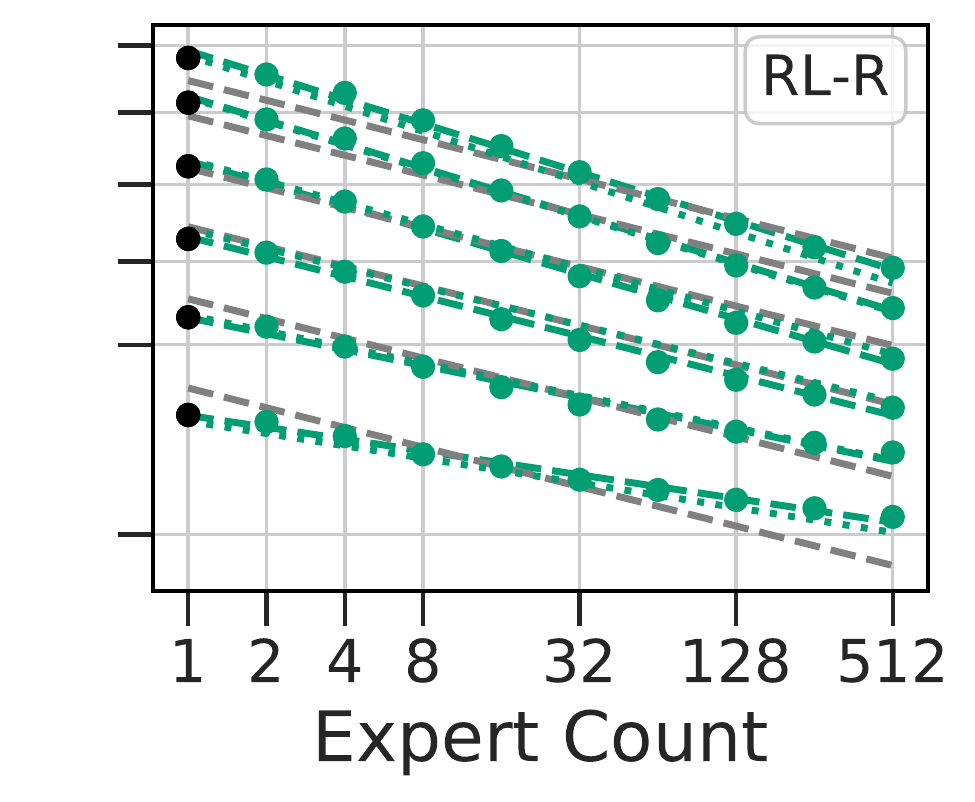}
  \hspace{-0.3em}%
  \includegraphics[width=0.3\linewidth]{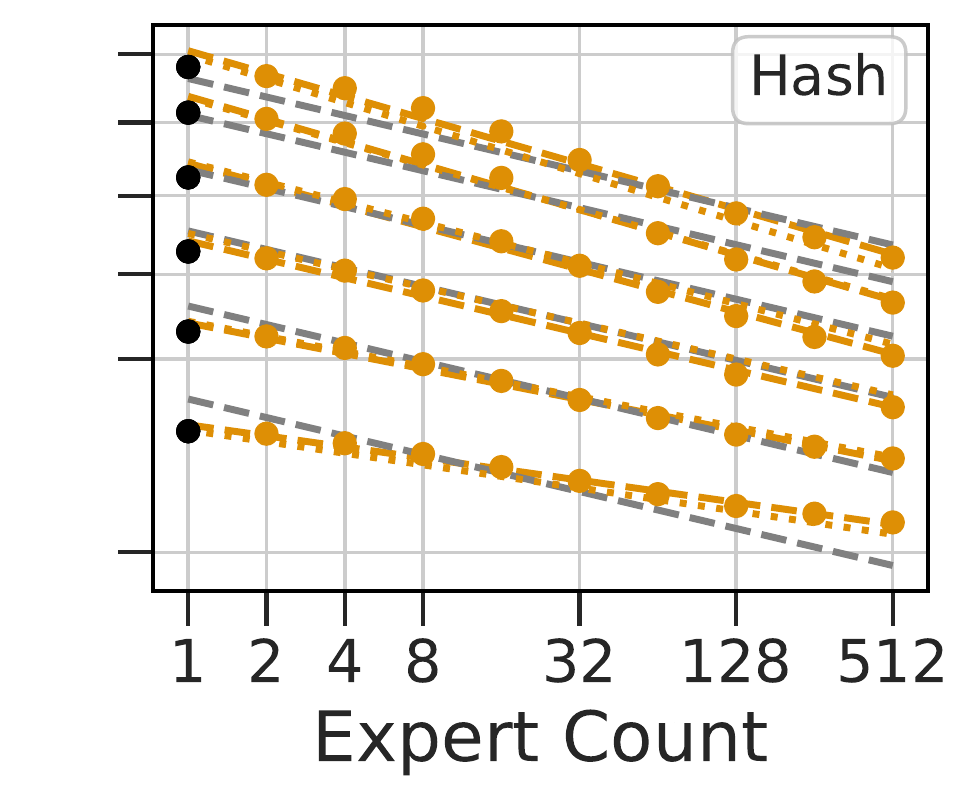}
\caption{Fits for \base, \rlr and \hash . Dashed lines are solutions to Eq.~\eqref{eq:hopeful_power} with $b(N)$ given by Table \ref{tab:alphas} while dotted lines are solutions to Eq.\eqref{eq:no_tailoff}. Solutions for Eq.~\eqref{eq:separable} are in grey. The separable solution fails to account for decreasing performance given by expert scaling.}\label{fig:linear_fits}
\end{figure}

This motivates us to introduce a simple extension: that of a multiplicative interaction between $\log N$ and $\log E$. This is conveniently the exact function which leads to $b$ scaling with $\log N$ and takes the following form:
\begin{gather}
% \begin{split}
       \!\!\!\log L(N, E) {\triangleq} {+} a \log N {+} b  \log E {+}
       c \log N \log E {+} d\label{eq:no_tailoff}
% \end{split}
\end{gather}
This function has the property that the log-log slope in both $N$ and $E$ are affine in the logarithm of the other variable. In other words, with $E$ or $N$ fixed, the performance $L$ scales with $N$ or $E$ following~\eqref{eq:dense_law} and \eqref{eq:hopeful_power} with slopes given by:
\begin{gather}
% \begin{split}
       a(E) \triangleq - \frac{\partial \log L}{\partial \log N} = a + c \log(E)\label{eq:a_e}\\
      b(N) \triangleq - \frac{\partial \log L}{\partial \log E} = b + c \log (N), \notag
\end{gather}
$b(N)$ matches the behavior reported in \autoref{tab:alphas}. A transposed table, fitting sets of models with fixed $E$ and changing $N$, can be found to match the behavior predicted by $a(E)$ (see \autoref{tab:all_alpha_ns}). There are two symmetric non-logarithmic representations of \eqref{eq:no_tailoff}, useful for comparison to \eqref{eq:multi_separable}:
\begin{align}
\stepcounter{equation}
       L(N, E) &= \left(\frac{10^{d/a}}{N}\right)^a\left(\frac{1}{E}\right)^{b + c \log (N)}\tag{\arabic{equation}a}\label{eq:factorized},\\[4pt]
       &= \left(\frac{10^{d/b}}{E}\right)^b\left(\frac{1}{N}\right)^{a + c\log(E)}.\tag{\arabic{equation}b}\label{eq:bad_factorized}
\end{align}
% % 

% %

% 
\begin{figure*}[t]
  \centering 
  \includegraphics[width=0.6\textwidth]{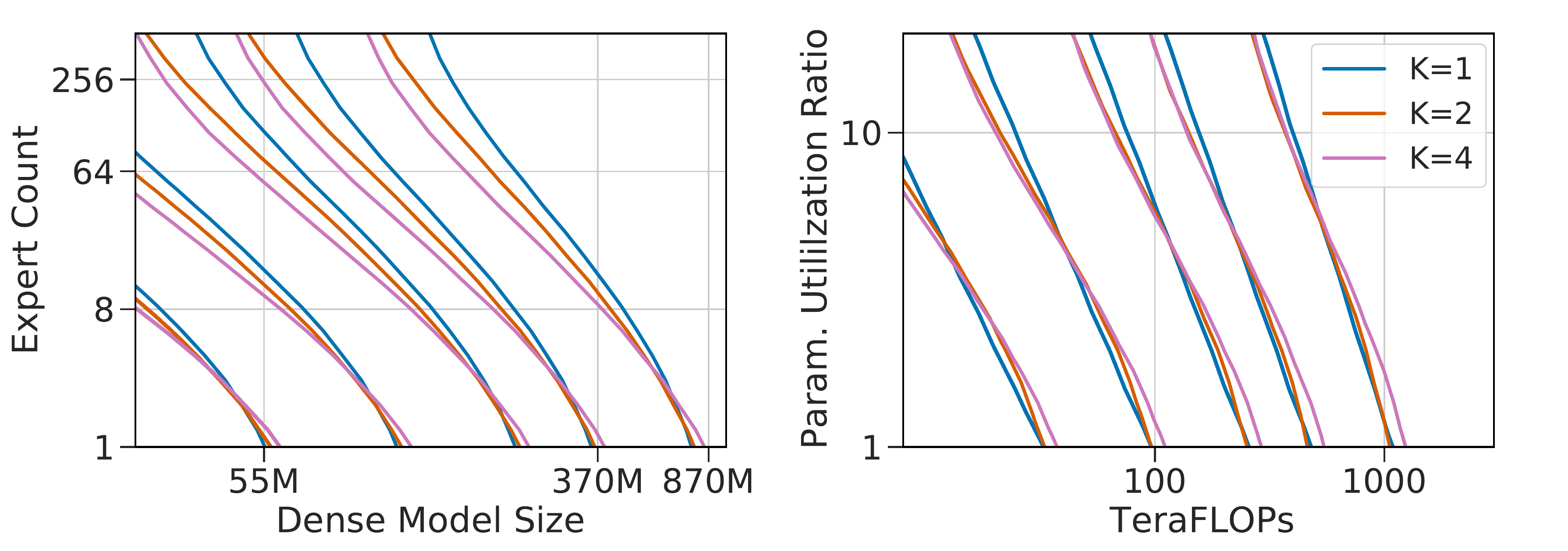}
  \includegraphics[width=0.6\textwidth]{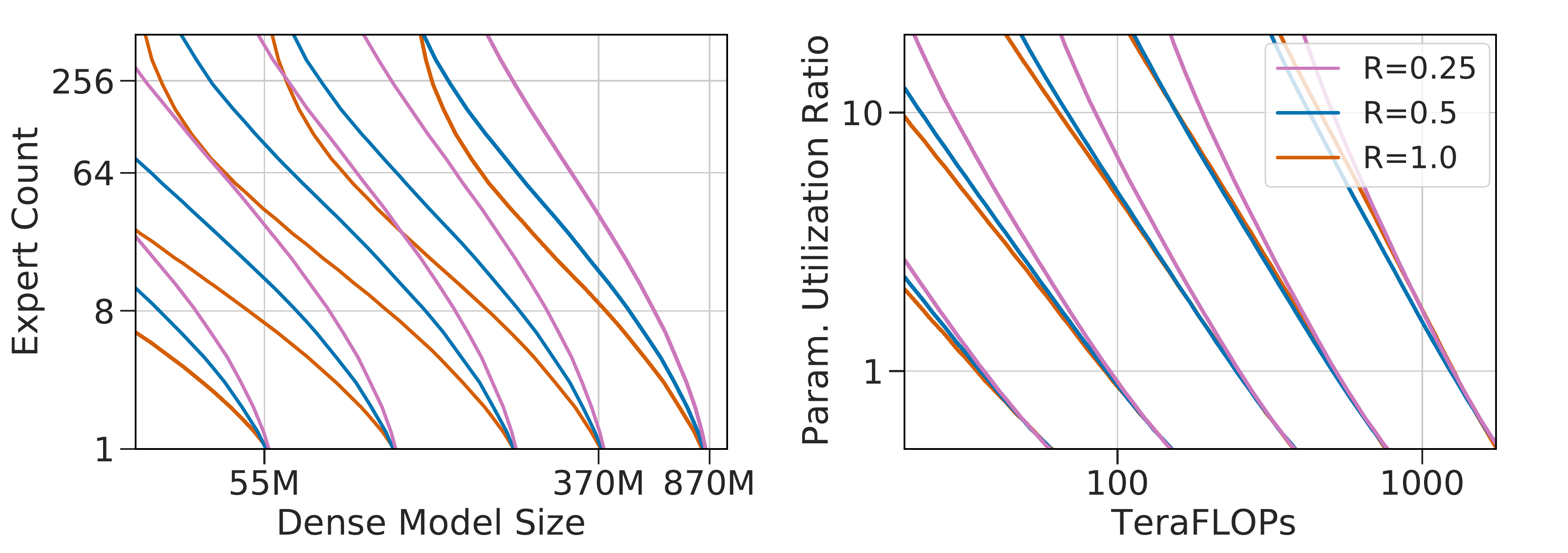}
\caption{Level curves for Equation \eqref{eq:real_joint_scaling_law} and Equation \eqref{eq:real_joint_scaling_law_fp} on \base for $K \in \{1, 2, 4\}$ (left two), $R \in \{1.0, 0.5, 0.25\}$ (right two). Scaling laws in $(N, E)$ differ for models with different values of $(K, R)$: indicated by non-overlapping level-curves. A change of variables to $(F, P)$ leads to almost-overlapping functions: allowing the same fits to be reused across changes in the routing architecture. 
}
\label{fig:ben_topk_fits}
\end{figure*}

\paragraph{Fitting.}
Fitting the bilinear~\eqref{eq:no_tailoff} instead of~\eqref{eq:separable} substantially reduces the prediction error for large~$N$ (\autoref{tab:rmse}, Eq.~\eqref{eq:separable} vs Eq.~\eqref{eq:no_tailoff}), as displayed in~\autoref{fig:linear_fits} (dotted lines match the dashed ones, where the grey separable fit doesn't). We verify dense equivalence: $\alpha_N \approx a$, while $N_c\approx \exp(d/a)$, and thus the law~\eqref{eq:no_tailoff} gives similar prediction to the reference law~\eqref{eq:dense_law} for dense models. Predictions for fixed $N$ are visualized as grey lines in \autoref{fig:joint_curves}.

\paragraph{Interpretation.}
In Eq.~\eqref{eq:no_tailoff}, when $c$ is positive, the expert improvement slope~$b(N)$ reduces with model size~$N$. All three routing techniques considered therefore predict \textit{diminishing improvements from routing when increasing scale}. However, the scaling of \base is predicted (and seen) to be substantially better. When designing a new technique, we can fit~\eqref{eq:no_tailoff} and predict a better scaling behavior if the fitted $c$ is lower than with other techniques. A clear goal for future work in routing techniques should be to find a method with scaling coefficient $c \approx 0$.

\subsection{Bounded Scaling in $E$}
\label{sec:tail_off}

Equation \eqref{eq:separable} models scaling in $E$ as a power law. For both small and large values of $E$, there are reasons to expect some deviation. If a routing technique degrades with $E$ (for instance, the variance of gradients in \rlr will increase), performance for large $E$ might be worse than predicted. On the other hand, fixed overhead (e.g., interference from auxiliary losses) might worsen scaling for low values of $E$, counter-intuitively leading to better than expected performance. Both phenomena appear clearly in \autoref{fig:joint_curves}. We seek to model this saturation such that the limit behavior in $E$ is bounded on both sides. We choose the following transformation, but discuss in~\autoref{sec:epc} a number of implications which are independent of the specific saturating form used:
\begin{align}
    \frac{1}{\widehat{E}} \triangleq
        \frac{1}{E - E_{\min} + \left(\frac{1}{E_\text{start}} - \frac{1}{E_{\max}}\right)^{-1}} + \frac{1}{E_{\max}}.\label{eq:saturation}
\end{align}
This is constructed
so that we have $\hat E(E_{\min}) = E_{\text{start}}$, while $\hat{E} \to  E_{\max}$ as $E \to \infty$. We fix $E_{\min} = 1$, indicating the lower bound of meaningful expert counts.
$\hat E$ can be seen as a thresholded version of $E$: increasing past $\Emax$ will give improvement, but not following a power law. Similarly, when $\Estart > 1$, $\hat{E} > E$ for small values of $E$. Practically, the fit is the same over a wide range of different thresholding functions.

\paragraph{Fitting.} Solving Equation~\eqref{eq:real_joint_scaling_law}, equal to Eq.~\eqref{eq:no_tailoff} with $E \to \hat{E}$, is complicated by its non-convexity.
We find the coefficients $(a, b, c, d, \Estart, \Emax)$ as the best of repeated solutions provided by the L-BFGS-B algorithm \citep{byrd1995limited}. \autoref{fig:joint_curves} shows fitted curves from these equations; coefficients are reported in~\autoref{tab:final_coeffcients}.

\paragraph{Interpretation.} Relative to using the simple bilinear law~\eqref{eq:no_tailoff}, fitting Eq.~\eqref{eq:real_joint_scaling_law} improves prediction for the lowest and highest values of $E$ considered. Crucially, while the deviation from a power-law (and therefore improvement in RMSLE) is relatively minor for the values of $E$ considered, the deviation is nonetheless clear (seen best looking at the raw losses in \autoref{fig:experts_and_params1}). We believe it is important to model this saturation because (as argued in \autoref{sec:n_max}) the limit behavior of model performance as $N$ increases is substantially different when bounded, with important properties that are independent of $\Emax$. We further hypothesize that future work, able to test still larger values of $E$, will see a more quantitative benefit from including these terms. 
This can be already observed in \autoref{fig:fig3_hash} when noting that the law~\eqref{eq:no_tailoff} does not over and under estimate the performance for $E = \{2, 4, 256, 512\}$ as it does in \autoref{fig:linear_fits}.
Level curves of Eq.~\eqref{eq:real_joint_scaling_law} enumerate the $\{(N, E)\}$ which are predicted to achieve fixed performance, as visualized in Fig~\ref{fig:main}(b). This demonstrates of the power of routing: a model with $N=5M$ and $E=128$ equals the performance of a model with $N=55M$ and $E=1$,which requires over ten times more compute per inference.

\subsection{Generalizing Across Architecture Variants}
\label{sec:generalizations}

The models trained so far use fixed choices for two key details of routing: the number of experts executed per-datapoint $K$ and the frequency of routed layers across depth $R$ (previously set at 1 and $0.5$, respectively). For any selected value of $K$ and $R$ we may fit Eq.~\eqref{eq:real_joint_scaling_law} to observed performance, but since these variables are independent of $N$ and $E$, we do not expect the same coefficients to remain valid across values of $K$ and $R$. To allow for a unified scaling law, we modify Eq.~\eqref{eq:real_joint_scaling_law} to use terms in $F$, the TeraFLOPs required per forward pass, and in the ratio $B \triangleq \frac{P}{F}$ where $P$ is the total number of parameters. Specifically, $F$ is motivated by the approximation from \citet{kaplan2020scaling} that $F = 2N$. $B$, the \textit{parameter utilization ratio}, is an affine function of $E$, close to linear when most parameters lie in the routed components of the model. 

Using $(F, B)$ instead of $(N, E)$ (and setting $E_{\min}$ to $\frac{1}{2}$) results in Eq.~\eqref{eq:real_joint_scaling_law_fp}.
To show the advantage of this change of variables we conduct two experiments: varying $K$ across $\{1, 2, 4\}$ and $R$ across $\{0.25, 0.5, 1.0\}$. In both cases, we vary $E \in \{8, 64, 256\}$ and $N \in \{15M, 370M, 870M\}$.

\paragraph{Fitting.} Eq.~\eqref{eq:real_joint_scaling_law_fp} predicts the scaling behavior of models as well as Eq.~\eqref{eq:real_joint_scaling_law} for a given routing architecture, as indicated in~\autoref{apfig:pf_and_expert_fits}.
The benefit of the change of variables is seen most clearly in \autoref{fig:ben_topk_fits}, which plots contours of fixed loss value as functions of $(N, E)$ and of $(F, B)$. For varying $(K, R)$, the loss surface as a function of $N$ and $E$ changes: meaning a joint fit would be inaccurate. Plotted as functions of $(F, B)$, the loss surface is almost the same, suggesting a shared fit between all three methods (see \autoref{apfig:pf_fits_topk_individual} and \autoref{apfig:pf_fits_ben_individual} for joint fits for $K$ and $R$ respectively). We highlight that $R = 0.25$ deviates slightly. Plausible explanations are discussed in \autoref{sssec:route_one_layer}. The possibility to use a shared fit indicates a singular takeaway: the architectural details $K$ and $R$ little affect the scaling behavior of a Routing Network. The loss of the network can thus be predicted based only on inference flops $F$ and total number of parameters $P$.

\section{Scaling Law Applications}\label{sec:applications}

Next we provide two applications of the scaling laws presented. We re-emphasize that all values are only valid at the specific token count all models were trained at: 130B. \autoref{apsec:convergence} provides evidence that our analysis, if not the numerical values, are nevertheless robust to token count.

\subsection{Effective Parameter Equivalence}
\label{sec:epc}

We leverage Eq.~\eqref{eq:real_joint_scaling_law} to compute the size~$\bar N$ of a dense model giving the same performance as a Routing Network. Specifically, we solve for $L(\bar{N}, 1) = L(N, E)$, yielding
\begin{equation}
    \label{eq:equivalence_tail_off}
    \bar{N} \triangleq
    {\left(N\right)}^{\alpha(\hat E) / \alpha(\Estart)}
    {\left(\hat{E} / \Estart\right)}^{b / \alpha(\Estart)}
\end{equation}
Here $\alpha(E) = a + c \log E$. Given a model with $N$ and $E$, we call $\bar{N}$ that model's \textit{Effective Parameter Count} (or \epc). Eq.~\eqref{eq:real_joint_scaling_law} predicts that the performance of all models increases as a power law in this variable
\begin{equation}
    \log L(N, E) = a \log \bar{N}(N, E) + d.
\end{equation}
The result of plotting all models as a function of $\bar N$ is shown in \autoref{fig:main}(c): a good fit across four orders of magnitude. Scaling in terms of $\bar{N}$ results in a unifying power law: valid for dense and routed language models alike.

\subsection{Routing Behavior for Large $N$}
\label{sec:n_max}

\begin{table}[t]
% \small
\centering
\caption{Solutions to Eq.~\eqref{eq:real_joint_scaling_law}.}
\begin{tabular}{ c | c c c c c c}
\toprule
 & a & b & c & d & $E_\text{start}$ & $E_{\max}$ \\
\midrule

\textbf{\base} & -0.082 & -0.108 & 0.009 & 1.104 & 1.847 & 314.478 \\
\textbf{\rlr} & -0.083 & -0.126 & 0.012 & 1.111 & 1.880 & 469.982 \\
\textbf{\hash} & -0.087 & -0.136 & 0.012 & 1.157 & 4.175 & 477.741 \\
\bottomrule

\end{tabular}
\label{tab:final_coeffcients}
\end{table}

\begin{figure}[t]
    \centering
    \includegraphics[width=.56\linewidth]{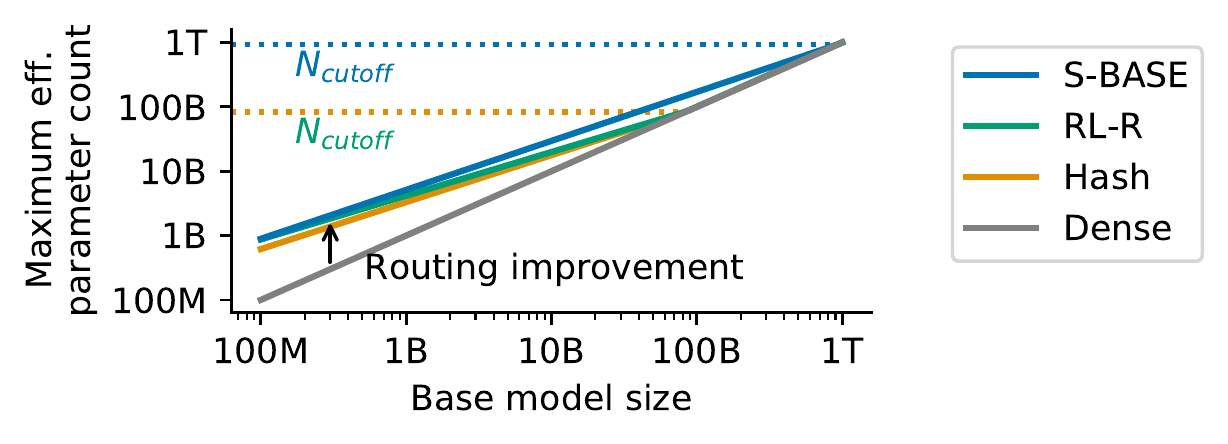}
    \caption{Maximum effective parameter count as a function of base model size. Routing helps until a certain size $\Ncut$, that varies strongly between methods (\base being the best)}
    \label{fig:epc_max}
\end{figure}

\epc leads to a better grasp of the behavior of routing as $N$ increases. Of immediate interest is $\Ncut$: the value of $N$ where $\bar{N}(N, E) \leq N$. For larger $N$, routing will not improve performance. This is easily found to obey $\log \Ncut = \frac{b}{c}$. $\Ncut$ equals $937\textrm{B}$, $85\textrm{B}$ and $83\textrm{B}$ for \base, \rlr and \hash respectively. These values are highly dependent on the number of tokens seen, and $\Ncut$ is expected to increase with increased numbers of tokens.

Next we consider $\bar N_{\max}(N) \triangleq \max_E \bar N(N, E)$, i.e. the maximal effective parameter count that a routing network can reach. Eq.~\eqref{eq:equivalence_tail_off} predicts that $\log \bar N$ is an affine function of $\log N$ for any fixed $E$, and $\bar N_{\max}(N) = N$ for $N > \Ncut$. Therefore $\log {\bar N}_{\max}$ is piecewise-affine in $\log N$, as displayed in \autoref{fig:epc_max}:
\begin{align}
    \foralls N \leq \Ncut = 10^{-\frac{b}{c}},\quad \bar N_{\max}(N) &= \bar N(N, \Emax),\notag \\
    \forall N \geq \Ncut, \bar N_{\max}(N) &= N.
\end{align}
Note that $\bar{N}_{\max}$ is continuous near $\Ncut$, since for all $E$, $\bar N(\Ncut, E) = \Ncut$. Moreover, the slope of $\bar N_{\max}(\cdot)$ for $N \leq \Ncut$ is positive whenever $\Emax \leq \Estart 10^{-a/c}$, which is true for our coefficients. In this setting $\bar N_{\max}(\cdot)$ is a non-decreasing function of $N$.
%---this reasonable characteristic is imputable to using the saturation transform~\eqref{eq:saturation}.
Therefore for any routing network where $N < \Ncut$, $N \leq \bar N_{\max}(N) \leq \Ncut$, meaning routing will never let you train a model more powerful than $\Ncut$. Note that despite this value not depending on $\Emax$, its existence crucially depends on the saturating transformation: without it ${\bar N}_{\max}$ is unbounded.

\subsection{Comparative Analysis}
\label{sec:comparisons}

\citet{kaplan2020scaling} use scaling laws to encapsulate and contrast the behavior of entire model classes. Here we mirror this analysis by using the scaling laws we have proposed to summarize the relative behavior of the three routing techniques considered. We make four concrete observations:

\begin{itemize}[itemsep=3pt]

    \item \base consistently outperforms \rlr and \hash, though \rlr is very competitive at smaller $N$.
    
    \item All routing techniques suffer from reducing efficacy as $N$ increases. Amongst the three techniques, \base scales best: the fitted parameter $c$ is lowest.
    
    \item For small $N$, \rlr and \base scale similarly with expert count and better than \hash (as indicated by computing the effective expert slope $b(N) = b + c \log N$).
    
    \item
    \hash and \rlr maintain power-law behavior for longer than \base (larger $\Emax$). However they suffer from more interference ($c$); leading to worse performance for most model sizes.
    
    \item
    \hash has large initial overhead (bigger $E_\text{start}$), clearly visible as a more obvious curvature at small $E$.

\end{itemize}

For a practitioner interested in applying routing techniques, we conclude with some recommendations:

\begin{enumerate}[itemsep=0pt,topsep=0pt]
  \item Use routing when training any model with $N \leq \text{1.3B}$.
  
  \item \base is a good default routing algorithm. \rlr will sometimes match \base in performance but is less robust and scalable (\autoref{sec:sensitivity}).
  
  \item Target using $E \in \{64, 128\}$ experts. Larger values will continue to improve, but with diminishing returns.
  
  \item Use $K{=}1$ experts. Route layers at frequency $0.5 \leq R \leq 1$; lower frequency reduces performance.
  
  \item Future routing research should focus on the terms $c$ and $\Emax$; indicative of limits to arbitrary scaling.
  
  \item New routing techniques must be validated at multiple values of $N$ and $E$ when comparing with prior work. Results on single sizes cannot be extrapolated.

\end{enumerate}

\section{Related Work}

In studying the empirical aspects of scaling, this work follows \citet{kaplan2020scaling}; which triggered much research including \citet{henighan2020scaling}, \citet{hernandez2021scaling} and \citet{ ghorbani2021scaling}. The underlying theory is less understood, but there is some exploration of this space including \citet{hutter2021learning} and \citet{bahri2021explaining}.

These studies, and ours, are mutually reliant on a large corpus of work improving the scalability of Transformers. This includes models like GPT-2 \citep{radford2019language}, GPT-3 \citep{brown2020language}, Jurassic-1 \citep{lieber2021jurassic} and Gopher \citep{rae2021scaling}, as well as work improving the ability of these models to be efficiently parallelized across multiple devices, including \citet{shoeybi2019megatron}, \citet{harlap2018pipedream}, \citet{kim2021scalable} and \citet{xu2021gspmd}.

Parallel to all this has been a long study of Routing Networks; a term introduced by \citet{rosenbaum2018routing} but developed extensively in the literature as Conditional Computation \citep{bengio2013estimating,bengio2015conditional,bengio2017reinforcement,denoyer2014deep} and Mixture of Experts~\citep{jacobs1991adaptive,collobert2003scaling,eigen2013learning}. The framework is sometimes further generalized, seen as per-example architecture search in \citet{ramachandran2018diversity} or as a graph problem in \citet{denoyer2014deep}. Routing was popularized for large scale training by \citet{shazeer2017outrageously}, and furthered by work including GShard \citep{lepikhin2020gshard}, Switch Transformer \citep{fedus2021switch} and GLaM \citep{du2021glam}. In this vein, \citet{artetxe2021efficient} undertake a comparative analysis of dense networks and \smoes with $E = 512$ that aligns with our results. Finally, the core routing architecture is still being improved.
\citet{nie2021dense} adapt $K$ through training where \citet{hazimeh2021dselect} learn it via a differentiable loss. \citet{ramachandran2018diversity} increase $K$ through depth and encourage architectural diversity across experts. \citet{caccia2021anytime} grows $E$ throughout training and \citet{rajbhandari2022deepspeedmoe} propose networks where $E$ changes with depth.

\section{Conclusion}

Using conditional computation to scale neural networks has long been a research goal, and methods based on Routing Networks have been increasing in popularity. Here we have introduced a scaling law (Eq.~\eqref{eq:real_joint_scaling_law}) that models the behavior of these networks. This scaling law predicts that, for all models considered, introducing routing into a language model improves performance. That improvement follows a power-law in the number of experts $E$ that diminishes with model size $N$,
and can be further generalized across routing architectures with Eq.~\eqref{eq:real_joint_scaling_law_fp}.
These scaling laws quantify the differences between three different routing techniques and lead to a single scalar (Eq.~\eqref{eq:equivalence_tail_off}) that simultaneously describes the performance of routed and dense models alike.

This work provides an empirical framework with which to analyze future innovations in routing. We hope the overwhelming evidence we provide towards the benefits of routing encourage it to be more rapidly adopted as a powerful tool for model improvement, whose scaling characteristics align with traditional methods of scaling (in depth and width) and which will remain beneficial up to models with base model size greater than 900 billion parameters.

\pagebreak

\section*{Acknowledgments}
We would like to thank Marc'Aurelio Ranzato, Nando de Freitas, Jacob Menick and Andy Brock for useful comments and feedback on early drafts of this paper. The infrastructure needed to train these models wouldn't have been possible without the dedicated work of the JAX and XLA teams, especially Peter Hawkins, Roy Frostig and James Bradbury who all were crucial in the development of the routing software.

%Bibliography
\bibliographystyle{plainnat}
\bibliography{arxiv_version}

\appendix

\section{Architecture}
\addcontentsline{sections}{section}{added via addcontentsline}
\label{app:archtecture}

Our Transformer \citep{vaswani2017attention} is based on the architecture in \citep{radford2019language} with relative positional encodings \citep{dai2019transformer}. Text is tokenized via SentencePiece \citep{kudo2018sentencepiece} with $32,000$ tokens and a byte-level backoff. We use Megatron-style FFW sharding \citep{shoeybi2019megatron} where useful. Parameters are stored in bfloat16 but all optimizer statistics are kept in float32. As a result, the activations of the language models are calculated in bfloat16 (though we explicitly upcast to perform all operations involving a softmax, including the Attention Block and Router, in full float32 precision). This is crucial to maintain stability on larger models \citep{fedus2021switch, rae2021scaling}. The learning rate starts at 1e-7 and decays to 2e-5 with a cosine decay rate over the entire $250,000$ steps, after an initial warmup phase ramping up to 2e-4 in the first $1500$ steps.

We use seven different model sizes, with names and architectures specified in the following table. The width of the hidden layer $d_{\mathit{ffw}}$ is fixed at four times the width of the activations $d_{\mathit{model}}$, and we use the same dimension for keys and values.

\begin{table}[h]
\begin{adjustbox}{width=0.5\columnwidth,center}

\begin{tabular}{ c | c c c c c}
 \toprule
  \textbf{Name}
  & $d_{\textit{model}}$
  & $n_\mathit{layers}$
  & $n_\mathit{heads}$
  & K/V size
  & Actual \# Params
  \\ [0.5ex] 
  \midrule
  15M & 512 & 6 & 8 & 32 & $16,527,360$ \\ 
  25M & 512 & 8 & 8 & 64 & $27,279,360$ \\ 
  55M & 640 & 10 & 12 & 64 & $57,369,600$ \\ 
  130M & 896 & 12 & 16 & 64 & $132,163,584$ \\ 
  370M & 1536 & 12 & 12 & 128 & $368,123,904$ \\ 
  870M & 2048 & 16 & 16 & 128 & $872,546,304$ \\ 
  1.3B & 2048 & 24 & 16 & 128 & $1,308,819,456$ \\ 
  \bottomrule
\end{tabular}
\end{adjustbox}
\caption{Model definitions used throughout this work.}
\label{tab:architectures}
\end{table}

\begin{figure*}[t]
  \centering
  \includegraphics[ width=0.9\textwidth]{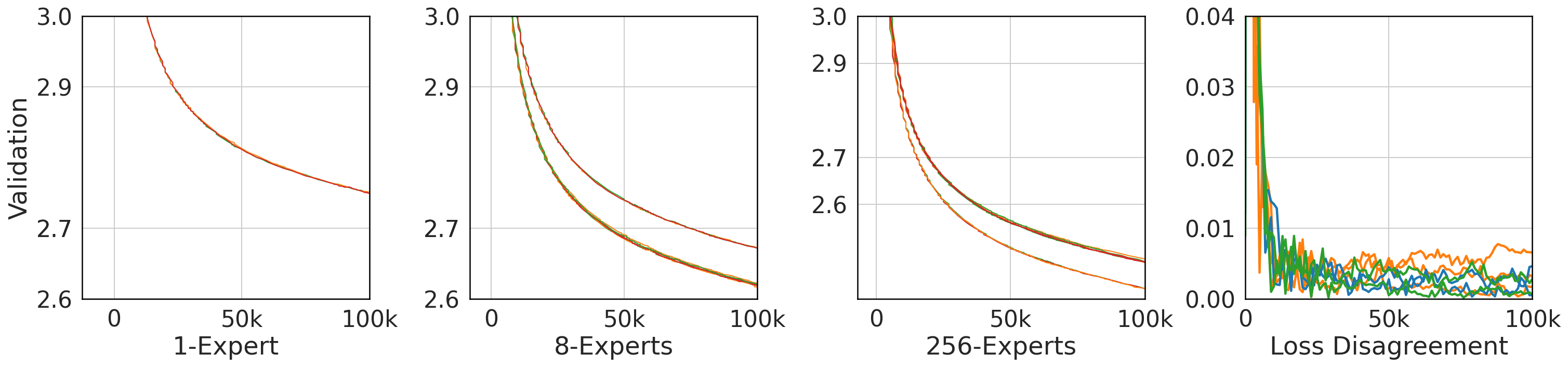}
\caption{Training curves color-coded by random seed for 1, 8 and 256 experts and the step-wise maximum disagreement between runs.}

\label{apfig:seeds}
\end{figure*}

The number of models we trained was too large to practically include multiple runs of each model with different seeds. To give an idea of the potential error introduced by random chance, we trained all three routing techniques with 3 different seeds on a 130M model for $100,000$ steps with $8$ and $256$ experts (along with a dense baseline). Results are shown in \autoref{apfig:seeds}. Different seeds (which influence not only parameter initialization but Expert Parallelism -- see Appendix \ref{apsec:routing}) lead to extremely minimal model divergence after an initial transitory period, with different seeds diverging by no more than $0.01$ before $100,000$ steps. This is a close match to the $0.02$ error mentioned in \citep{kaplan2020scaling}\footnote{Anecdotally, throughout the development of this work we used $0.02$ as the cutoff to denote statistical significance.}.

\section{Detailed Routing Techniques} 
\label{apsec:variants}

Here we detail aspects of the routing techniques crucial to their implementation and provide comparisons to key alternatives.

\subsection{Balancing Losses}

We encourage uniform routing in both our SMoE and RL-R methods with the differentiable load balancing loss adapted from the mean square auxiliary loss in \citet{shazeer2017outrageously} and introduced in \citet{lepikhin2020gshard, fedus2021switch}.
\begin{equation}
    L_B = {E} \cdot \sum_{e=1}^{E} m_e \cdot \frac{g_e}{N}
    \label{eq:moe_loss}
\end{equation} 
Where $m_e$ is the mean gate per expert:
\begin{equation}
    m_e = \frac{1}{N} \sum_{x \in B} p_e(x)
\end{equation}
And $g_e$ is the gating decision per expert:
\begin{equation}
    g_e = \sum_{x \in B} 1 \{\argmax  p(x), e\}
\end{equation}
For $x$ in batch $B$ of size $N$ and policy $p(x) = \softmax(W_p x + b_p)$. There are two cases where the selected experts may not be the ones used: in \base after the Sinkhorn redistribution step (see \autoref{sec:sinkhorn}) and when experts are skipped due to load-balancing (see \autoref{sec:redistribution}). In both cases, the balancing loss is applied to the original gating decisions made by the policy. We found that the auxiliary loss is less effective if post-balancing experts were considered.

\subsection{SMoE with Sinkhorn redistribution (\base)}

Our implementation of \base differs from that proposed in \citet{lewis2021base} in two ways. First, we replace the auction algorithm for re-assigning expect selections with a continuous rebalancing process implemented via a Sinkhorn algorithm \citep{cuturi2013sinkhorn, gabriel2019ott}. Second, we add a shuffling step, similar to \citet{lewis2021base}, before computing the optimal assignment via Sinkhorn per-device (as opposed to across all devices as done in \citet{lewis2021base}). In addition, we did not use any input jitter on the activations sent to $\rho$ as we did not see a noticeable effect. This is in line with \baseorig but differs from recommendations in other \smoe papers \citep{lepikhin2020gshard, fedus2021switch}.

\subsubsection{Sinkhorn Redistribution} \label{sec:sinkhorn}

We rebalance expert selections using a Sinkhorn layer applied on top of the router logits, an idea that was explored independently in parallel by \citet{DBLP:journals/corr/abs-2109-11817}. This is substantially more efficient on our accelerator cluster than a hard matching algorithm. We consider $H \in \RR^{T \times d}$ the intermediary embeddings of the networks before the application of a routed layer (folded on the batch and time axes of respective sizes $b$ and $t$, with $T \triangleq b t$). Those are fed to the linear router, which output a logits matrix $L_i = H_i W + b \in \RR^{T \times e}$. Here $E$ is the number of experts, and $W \in \RR^{d \times E}$ and $b \in \RR^{E}$ are the router parameters. From these logits, \smoe and \rlr computes expert selection probabilities $\Pi$ by applying a softmax operation along the expert axis. In doing this, we compute selection probabilities for each input separately, without taking into consideration any capacity constraints on expert, forcing us to introduce load-balancing later (\autoref{sec:redistribution}). We seek a proper way to integrate constraints in a mathematically grounded framework. 

Mathematically, $\Pi$ is obtained by solving a simple problem with constraints: each input must, on average, prefer exactly one expert. This is made clear by the variational formulation of the softmax:
\begin{equation}
    \Pi \in \RR^{T \times E} \triangleq [\softmax(L_i)]_{i \in [1, T]} \\
    = \argmax_{\substack{
    \Pi \geq 0, \\
    \foralls i \in [T], \sum_{j \in [E]} p_{ij} = 1,
    }}
    \dotp{\Pi}{L} - H(\Pi)
\end{equation}
where $H$ is the Shannon entropy of the matrix $\Pi$, i.e. $H(\Pi) \triangleq \sum_{i =1}^T \sum_{j=1}^E p_{ij} \log p_{ij}$, and $[ \cdot ]$ denotes horizontal stacking. This variational formulation offers a natural alternative to incorporate extra constraints. For ideal performance, each expert should be assigned the same number of tokens on average $B = \frac{T}{E}$. We therefore add $E$ additional constraints: 
\begin{equation}
    \Big\{ \forall\, j \in [E],\, \sum_{i = 1}^T p_{ij} = B \Big\},
\end{equation}
which yields the doubly constrained regularized linear problem
\begin{align}
    & \Pi \in \RR^{T \times E} \triangleq \argmax \dotp{\Pi}{L} - H(\Pi),\label{eq:sinkhorn} \\
    & \text{under the constraints}\qquad
    \left\{
    \begin{array}{l}
        \Pi \geq 0, \\
    \foralls i \in [T], \sum_{j=1}^E p_{ij} = \frac{1}{T}, \\
    \foralls j \in [E], \sum_{i=1}^T p_{ij} = \frac{1}{E}
    \end{array}
    \right. \notag
\end{align}
that we recognize as the regularized Kantorovich problem of optimal transport \citep{kantorovitch_translocation_1958,cuturi2013sinkhorn}.

We solve this problem using the Sinkhorn algorithm \citep{knopp_concerning_1967}, that takes the logit matrix $L \in \RR^{T \times E}$ and returns a soft-assignment matrix $\Pi \in \RR^{T \times E}$. The Sinkhorn algorithm solves Eq.~\eqref{eq:sinkhorn} by alternated ascent in the dual (see~\citet{gabriel2019ott} for details). Starting from $f_0 = 0 \in \RR^T$ and $g_0 = 0 \in \RR^E$, we set
\begin{align}\label{eq:sinkhorn_update}
\foralls i \in [T],\qquad {(f_{t+1})}_i &= - \log \frac{1}{E} \sum_{j=1}^E \exp(L_{ij} - {(g_t)}_j), \\
\foralls j \in [E],\qquad {(g_{t+1})}_j &= - \log \frac{1}{T} \sum_{i=1}^T \exp(L_{ij} - {(f_{t+1})}_i).\notag
\end{align}
These updates converge towards an optimal couple $(f, g)$, such that
\begin{equation}
    \Pi = \frac{1}{T E} \exp(L + f \oplus g)
\end{equation}
is the solution to Eq.~\eqref{eq:sinkhorn}, where $(f \oplus g)_{ij} \triangleq f_i + g_j$ for all $i, j \in [T] \times [E]$. As detailed below, we early stop the iterations~\eqref{eq:sinkhorn_update} by measuring the primal violation of constraints in $L_1$ norm, i.e. when
\begin{equation}\label{eq:stopping_criteria}
    \sum_{j=1}^E \left\vert \sum_{i=1}^T {(\Pi_t)}_{ij} - \frac{1}{E} \right\vert
    + \sum_{i=1}^T \left|\sum_{j=1}^E {(\Pi_t)}_{ij} - \frac{1}{T}\right| \leq e_{\textrm{tol}}
\end{equation}

Once the plan is computed, we greedily select, for each token, the device with highest device-selection probability, effectively applying an $\argmax$ operation on top of the Sinkhorn logits to form a transportation plan projection.

\paragraph{Comparison to \base and performance.} Compared to using an exact (early-stopped) auction algorithm as \citet{lewis2021base}, the complexity of the Sinkhorn algorithm is in $\Oo(N \times E)$ versus $\Oo((N \times E)^{3/2})$, and its update are well adapted to batch computations on TPU/GPU. In contrast, the auction algorithm must be run on CPU as it is a greedy per-coordinate algorithm; it becomes a computational bottleneck applied to models with many routed layers. Replacing the softmax output by an regularized optimal transport plan is very naturally interpreted as adding a balancing distribution constraint to the softmax operator. Using an auction algorithm on top of the softmax assignment does not have this property.

Moreover, the Sinkhorn algorithm can be halted before it has fully converged with a proper tolerance parameter~\eqref{eq:stopping_criteria} where \citet{lewis2021base} uses a hard number of iterations. We find an error tolerance of $e_{\textrm{tol}}=10^{-2}$ gives consistently good performance. In practice we observe an end-to-end model overhead of $1\%$ to $3\%$ compared to Switch (the same routing technique without this reassignment). This computational offset is negligible compared to the per-step performance gain.
%shown in Figure \ref{fig:switch}
Without the rebalancing step, Switch is very sensitive to balancing loss hyperparameters (as noted in \citet{lewis2021base}) whereas \base maintains uniform routing decisions with improved performance and robustness while varying $E$ and $N$. 

\subsubsection{Shuffling Tokens}

Similar to \citet{lewis2021base}, we shuffle router inputs across workers by first computing a random permutation of the inputs and sending the $t$th row of the batch to the $\lfloor \frac{tE}{T} \rfloor$th worker. We found that this shuffling stage was necessary to prevent training from becoming unstable at larger scales. Our hypothesis is that the re-assignment provides a subtle side channel through which information can be propagated backwards in time, and this can be abused by larger models resulting in the validation loss diverging during training. Adding a shuffling stage ameliorates this issue by introducing a large number of irrelevant elements to the rebalancing process, making it harder to infer behavior of future inputs. Further work is needed to confirm this theory, but the introduction of the shuffling step does eliminate this performance degradation.

\subsection{Routing with Reinforcement Learning (\rlr)}
\label{apsec:rlr_variants}

\begin{figure*}[t]
  \centering
  \includegraphics[width=0.5\textwidth]{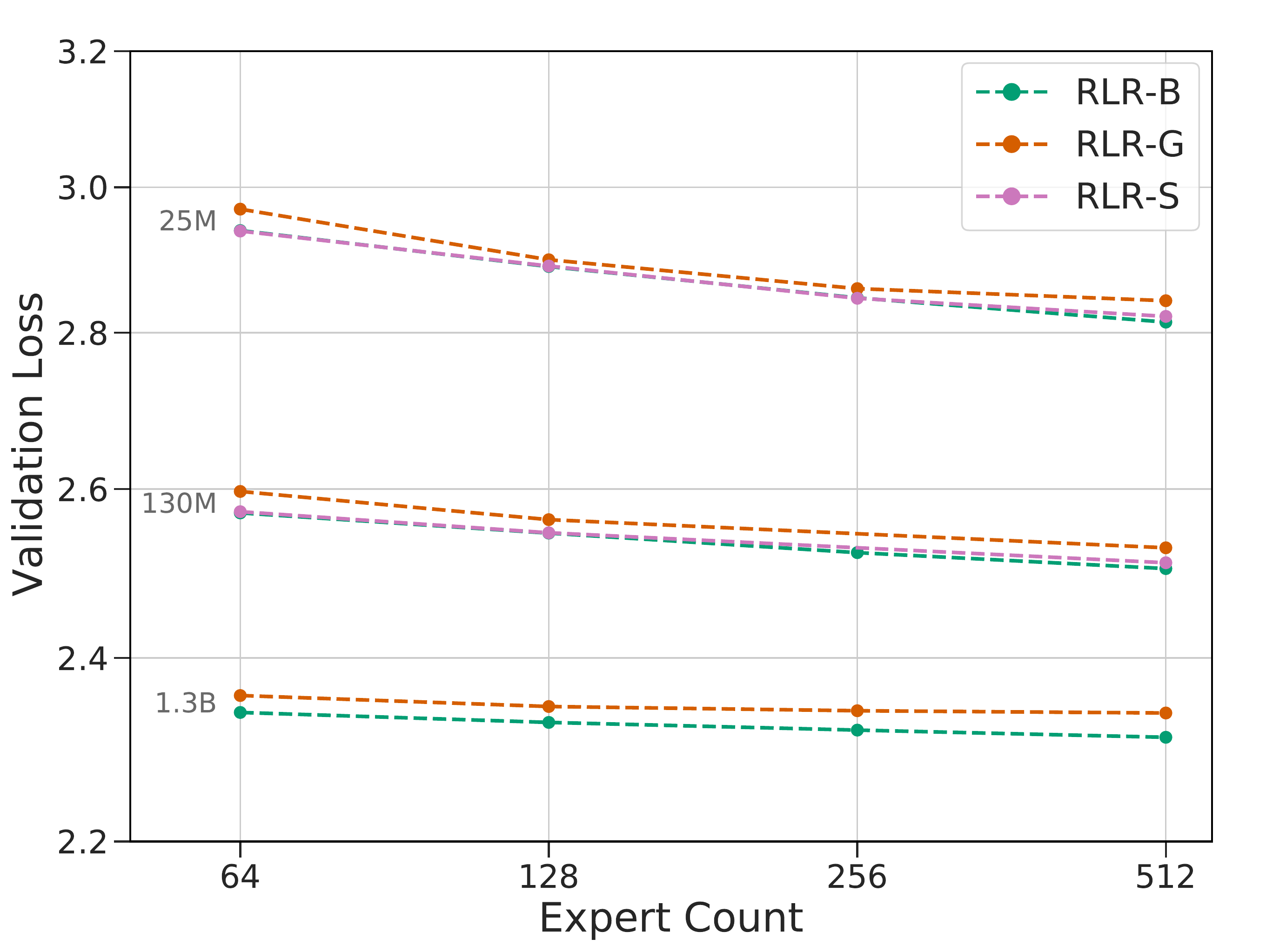}
\caption{The RLR-B method consistently outperforms RLR-G and RLR-S across scales. We found that Nucleus Sampling gives a significant improvement over Greedy Reinforce. However, performance is slightly improved by adding a learned baseline.}
\label{fig:basic_rl}
\end{figure*}

We will first describe a naive REINFORCE \citep{williams1992simple} implementation of routing, then describe possible extensions and improvements which lead to the form used in the main text as \rlr. 

Our implementation of REINFORCE uses the balancing loss in Equation \ref{eq:moe_loss} and a policy gradient loss:
\begin{equation}
    L =  \frac{1}{N} \sum_{i=1}^{N} \log \pi_i \cdot R_i
    \label{eq:basic_rl_loss}
\end{equation}
Where $R_i$ is the reward for each sequence in the batch of size N and $\pi$ is the normalized expert preferences output by a linear transformation as in \smoe. The proper thing is for $\rho$, the selected experts, to be samples from the distribution $\pi$, but we found that this substantially degraded performance at larger scales. This phenomenon can be attributed towards unwanted interference, where exploratory steps for $\rho$ which turn out to be unnecessary lead to bad gradient updates to the rest of the network \citep{rosenbaum2019routing}. We therefore consider a greedy selection method, where router outputs are selected as $\rho(x) = \textrm{TopK}(\textrm{softmax}(W_px + b_p))$. 

While sampling (even when tuning softmax temperature) decreased the performance of the model, we would nevertheless like to regain some of its exploratory power. To ameliorate this, we can use Nucleus Sampling \citep{holtzman2019nucleus}, which samples from the top-$p$ set of experts $E^{(p)}$.
\begin{equation}
    P'(e) = \begin{cases}
          P(e) / p' \quad &\text{if} \, e \in E^{(p)}, \\
          0 \quad &\text{otherwise.} \\
     \end{cases}
\end{equation}
Where $E^{(p)}$ is the smallest set of experts such that: 
\begin{equation}
    \sum_{e \in E^{(p)}} P(e) \geq p
\end{equation}
This eliminates the possibility of selecting experts with very low likelihood, while still introducing some randomness. It is important to emphasize that this introduces a distributional shift to the samples, which can be corrected with off-policy correction methods such as Importance Sampling.

An alternative improvement is to learn an additional baseline function for each router. This method has an additional entropy regularization loss and computes advantages $A_i=R_i-b_i$ for the learned baseline $b_i$:
\begin{equation}
    L = \frac{1}{N} \sum_{i=1}^{N} \log p_i \cdot A_i - \frac{1}{N} \sum_{i=1}^{N}  \log p_i \cdot p_i + \frac{1}{N} \sum_{i=1}^{N} v_i
    \label{eq:rl_loss}
\end{equation}
Where we use the Huber Loss to calculate the value loss $v_i$.
\begin{equation}
    v_i = \begin{cases}
          \frac{1}{2} (R_i - b_i)^2 \quad &\text{if} \, \abs{R_i - b_i} \leq \delta, \\
          \delta (\abs{R_i - b_i} - \frac{1}{2} \delta) \quad &\text{otherwise.} \\
     \end{cases}
\end{equation}
We numerate three \rlr variants below:
\begin{itemize}
    \item \textbf{Greedy REINFORCE (RLR-G)}. REINFORCE selecting the top-$k$ experts and no additional auxiliary losses.

    \item \textbf{Nucleus-sampled REINFORCE (RLR-S)}. REINFORCE using nucleus sampling to eliminate less reliable expert selections and reduce noise in the policy gradient update. In this method we sample from the top-$p$ truncated distribution. Nucleus sampling at a fixed top-$p$ scales well with increasing the number of experts.  

    \item \textbf{REINFORCE with baseline (RLR-B)}. Our RL method which stabilizes training with a learned baseline and a policy entropy regularization loss. We learn a baseline with a value function that has a single hidden layer of size $\frac{d_{\textit{model}}}{8}$.
\end{itemize}
Table \ref{tab:hyperparameters} details the hyperparameters chosen for each \rlr variant and \autoref{fig:basic_rl} contains validation losses across a number of models.
Note that the entropy loss is negative to encourage a more concentrated policy, and the weight must be tuned jointly with the load balancing loss to keep routing balanced. This is in line with \citet{bengio2015conditional}, who also use two loss terms to both encourage early specialization and expert diversity. Additionally, since the policy entropy loss has a similar effect to nucleus sampling, we did not see an improvement from including both regularization methods. RLR-B consistently performed the best, especially with regards to scalability in $E$ and $N$. For that reason we selected it as our prime example, and refer to it as \rlr elsewhere.

\begin{table}[ht]
\begin{center}
\caption{Selected hyperparameters for \rlr variants.}
\begin{tabular}{ c c c c}
 \hline
 Hyperparameter & RLR-G & RLR-S & RLR-B \\
 \hline
 Policy entropy weight & 0. & 0. & -5e-4 \\  
 Load balancing weight& 1. & 1. & 1.  \\  
 Policy gradient weight & 1e-1 & 1e-1 & 1e-2  \\ 
 Nucleus top-$p$ & - & 0.9 & 1.  \\  
 Value weight & - & - & 1e-2  \\ 
 Value hidden layers & - & - & 1  \\ 
 Value loss type & - & - & Huber  \\ 
\end{tabular}
\label{tab:hyperparameters}
\end{center}
\end{table}

\subsection{Hash layers (\hash)}
\label{sec:hash_details}

\hash is simple compared to \rlr or \base, but is highly reliant on the particular choice of hashing function. Many functions rely on knowing the integer ID which the tokenizer assigns to each unique token (characters, bytes, subwords, etc.). \citet{roller2021hash} describe multiple alternative functions, including pre-computing expert assignments for each token using a greedy assignment based on the frequency counts of the token on the training set. They do not observe any improvement in terms of perplexity relative to simpler random assignments of token to expert, but argue that balanced hashing has better properties for distributed training. 

\begin{figure*}[t]
\centering
\includegraphics[width=0.9\textwidth]{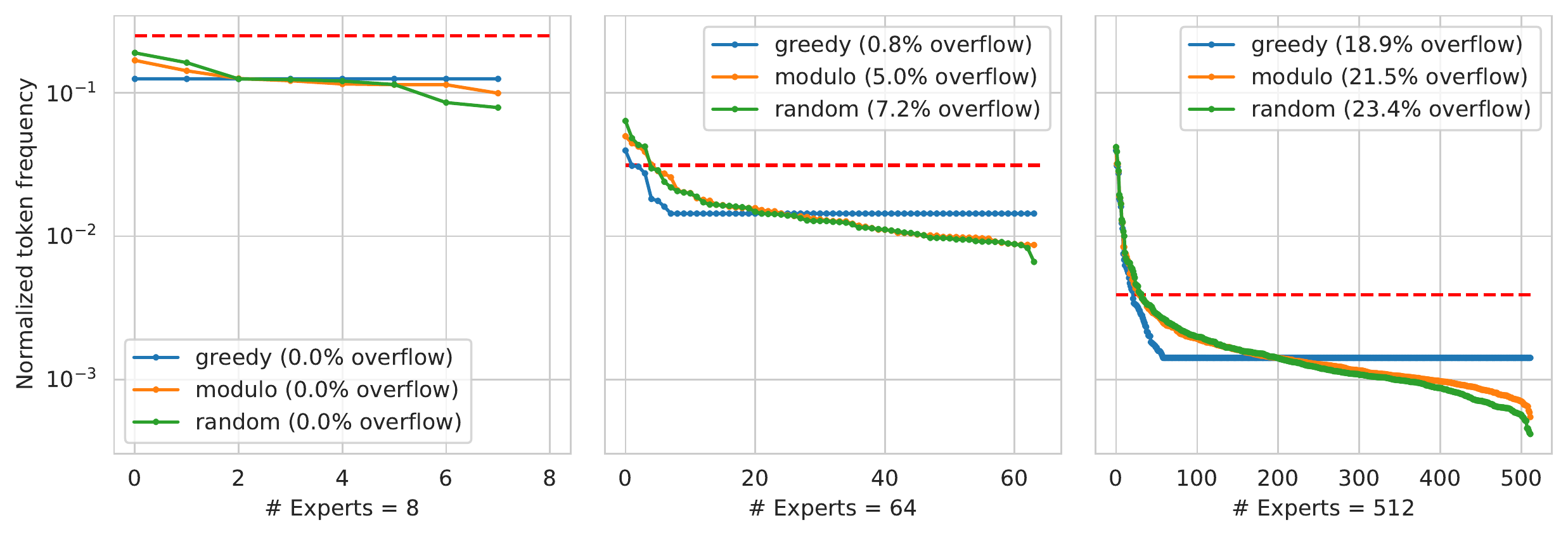}
\caption{
\hash becomes less balanced as $E$ increases. 
Here we compare three hash routing strategies using the token frequency in our validation set. The lines represent the amount of tokens sent to each expert, ordered from most subscribed to least subscribed.
The dotted line represents the point where tokens are likely to overflow under our bounded Expert Capacity setup ($C=2$).
\texttt{greedy} implements Balanced assignment as described in \citet{roller2021hash}, where the per-token frequency tables are pre-computed and tokens are assigned to the most empty expert ordered by frequency; \texttt{random} assigns each token to a random expert; and \texttt{modulo} uses the technique described in this paper. Note that (a) the token distribution is different from the one used by the tokenizer and (b) this simulation is based on marginal token frequencies, not batches of sequences. 
The \texttt{greedy} strategy does improve the workload for the mid range ($E=64)$, but not significantly for low ($E=8$) or high ($E=512$) numbers of experts. \texttt{modulo} provides a modest improvement over \texttt{random}.
}
\label{fig:hash-balancing}
\end{figure*}

Our implementation uses a simple modular hashing function, namely the token index modulo the number of experts. Tokens are indexed by our tokenizer in an order that is roughly ordered by their underlying frequencies in the training dataset, which means this strategy will be more balanced than an arbitrarily random assignment, while simpler to implement than fully balanced hashing. 
We note that poor balancing with increasing expert count is to some extent inevitable for any routing technique that defines one-to-one mappings between tokens and experts, assuming a bounded Expert Capacity (see Section~\ref{sec:redistribution}), as it becomes progressively harder to assign high frequency tokens into a bigger number of smaller buckets due to the tokens' heavy-tailed distribution. This can be seen in \autoref{fig:hash-balancing}.

\section{Distributed Routing Details}
\label{apsec:routing}

Here we describe the key aspects of Routing relevant to training on large clusters. We note there are several libraries available for supporting large-scale Routing, including DeepSpeed \citep{kim2021scalable, rajbhandari2022deepspeedmoe} and GSPMD \citep{xu2021gspmd}. Unfortunately these were incompatible with our preexisting infrastructure.

\subsection{Expert Parallelism}

We briefly review parallelism techniques, building up to Expert Parallelism, a technique for efficiently distributing parameters over an accelerator cluster. For a more in-depth exposition we recommend \citet{lewis2021base}, \citet{lepikhin2020gshard} or \citet{rajbhandari2022deepspeedmoe}. In a fully data-parallel world, every device has an identical copy of all parameters $\Theta$ and a different input batch $X$. Each device executes a forward and backward pass on $X$ and (usually) does a synchronous all-reduce across all devices on the gradients to $\Theta$. This is effective, but requires one copy of $\Theta$ for each device, wasteful when $|\Theta|$ is large.

The general class of techniques known as \textbf{Model Parallelism} reduce this duplication by having any individual device store only a subset of the entire model parameters. This reduction in memory comes with a cost: no longer can a single device take an input and produce the model's output; that device no longer contains all of $\Theta$. Most techniques therefore require some additional synchronization or data exchange.

\textit{Sharding Parallelism} \citep{shoeybi2019megatron} takes advantage of a mathematical property present both in 2-layer-MLPs and a Transformer's attention blocks: namely, that the output can be represented as the sum of $N$ components, where each component applies the same functional form with independent weights on the same input. \citet{shoeybi2019megatron} contains more details, but a simplified example can be given for a matrix multiplication where we observe the effect of splitting a matrix into columnwise sub-matrices: $Wx = [W_1, ..., W_N]x = \sum_i^NW_ix$. The effect of applying this technique such that each device has a separate subcolumn is to prevent the duplication of the weight matrices (which consist of the vast majority of $\Theta$). The disadvantage is that all devices must see the same input, meaning the total throughput of data on the cluster has been reduced $N$-fold. In addition, the sum described above is actually now a sum across devices, which introduces additional communication overhead.

\textit{Expert Parallelism} takes further advantage of the structure of a routed layer to similarly reduce the necessity of parameter duplication while avoiding the need to duplicate data between devices. In particular, rather than duplicating experts across all devices, each device contains only a subset of the experts which are not replicated anywhere else. Different devices still see different inputs. The key motivation is that a given input $x$ never needs to interact with the parameters corresponding to experts which the router did not send $x$ to. Therefore, a single input $x$ need only be present on a single device (the one which contains the experts which the router selected for $x$) to produce the correct output. In order to produce an output, the router selects an expert for all inputs and an additional data-exchange is introduced which sends all inputs to the device which contains the requested experts. Each device then processes the inputs it was sent, then returns all inputs to their original devices. Crucially, a roughly uniform router distribution leads to an evenly balanced computation across devices. This allows routed layers to be stored across a cluster with no duplicated data and without a reduction in data throughput. The downside is that this data exchange required across devices is generally more costly than the cross-device-sum required by sharding. More details are given in \citet{lewis2021base}. Previous work \citep{fedus2021switch} suggests using one expert per device. We believe this to be an implementation detail dependent on many aspects of the infrastructure in use. For us, typically using $4$ or $8$ local experts per device gave good performance.

All of Data, Sharding and Expert parallelism can be applied simultaneously. We use all three methods at will, selecting the combination which works fastest for a given cluster structure and model size. There are still more variations of model parallelism, notably \textit{Pipeline Parallelism} \citep{harlap2018pipedream, huang2019gpipe}, which we do not use.

\subsection{Load Balancing}
\label{sec:redistribution}

This at-will changing of parallelism techniques is dependent on the parallelism not affecting the output of the model. This is generally true, but expert parallelism brings in one complicating factor: \textit{load balancing}. In the description above, we emphasized that a roughly-uniform router (averaged over a minibatch) will send the same number of inputs to each device (we will call the expected value $BS_\text{avg}$). However, in the worst case all inputs on all devices might select the same expert, and therefore need to be sent to a single device. If memory is pre-allocated to accommodate this worse case, then each device must have enough free memory to potentially store the entire global batch size: prohibitive for large clusters.

The most common solution is to specify a \textit{capacity factor} $C$, and only allocate space for $BS_\text{avg} \times C$ tokens. When an expert is oversubscribed tokens are dropped at random until no experts are exceeding capacity. Having $C > 1$ is useful during training to prevent unnecessarily large numbers of tokens from being dropped. We set $C = 2$ for all experiments (though during evaluation we always allow all tokens to be routed to the desired expert). 
This strategy works well for the Transformer architecture due to its residual connections -- dropping a token means skipping that transformer block. As long as the amount of dropped tokens is kept at a reasonable bound, it does not impact learning.

An optimization we support is allowing an oversubscribed expert to use the memory allocated by an undersubscribed expert on the same device. This reduces the average number of tokens which are skipped, but does so at the minor cost of introducing an interaction between tokens being skipped and the specific co-habitation of experts on devices. In practice we do not find this to have a large effect. We note that the rebalancing used in \base substantially ameliorates the load balancing problem by attempting to force all experts to be assigned the same number of tokens. However because we use the approximate Sinkhorn algorithm, not a hard matching algorithm, over-subscription still happens (though at a much reduced rate) and so these steps are still taken.

\section{Architectural Variations} \label{sec:variations}

Throughout this work we have focused on a narrow subset of possible Routing Net architectures, which we believe are representative of recent work on large scale Routing Nets~\citep{roller2021hash,fedus2021switch,lewis2021base,shazeer2017outrageously,artetxe2021efficient,lepikhin2020gshard}. However, we also experimented with many variations of these architectures, some of which we highlight now in more depth.

\subsection{Robustness to hyper-parameter changes}
\label{sec:sensitivity}

\begin{figure*}[t]
  \centering
  \includegraphics[width=0.5\textwidth]{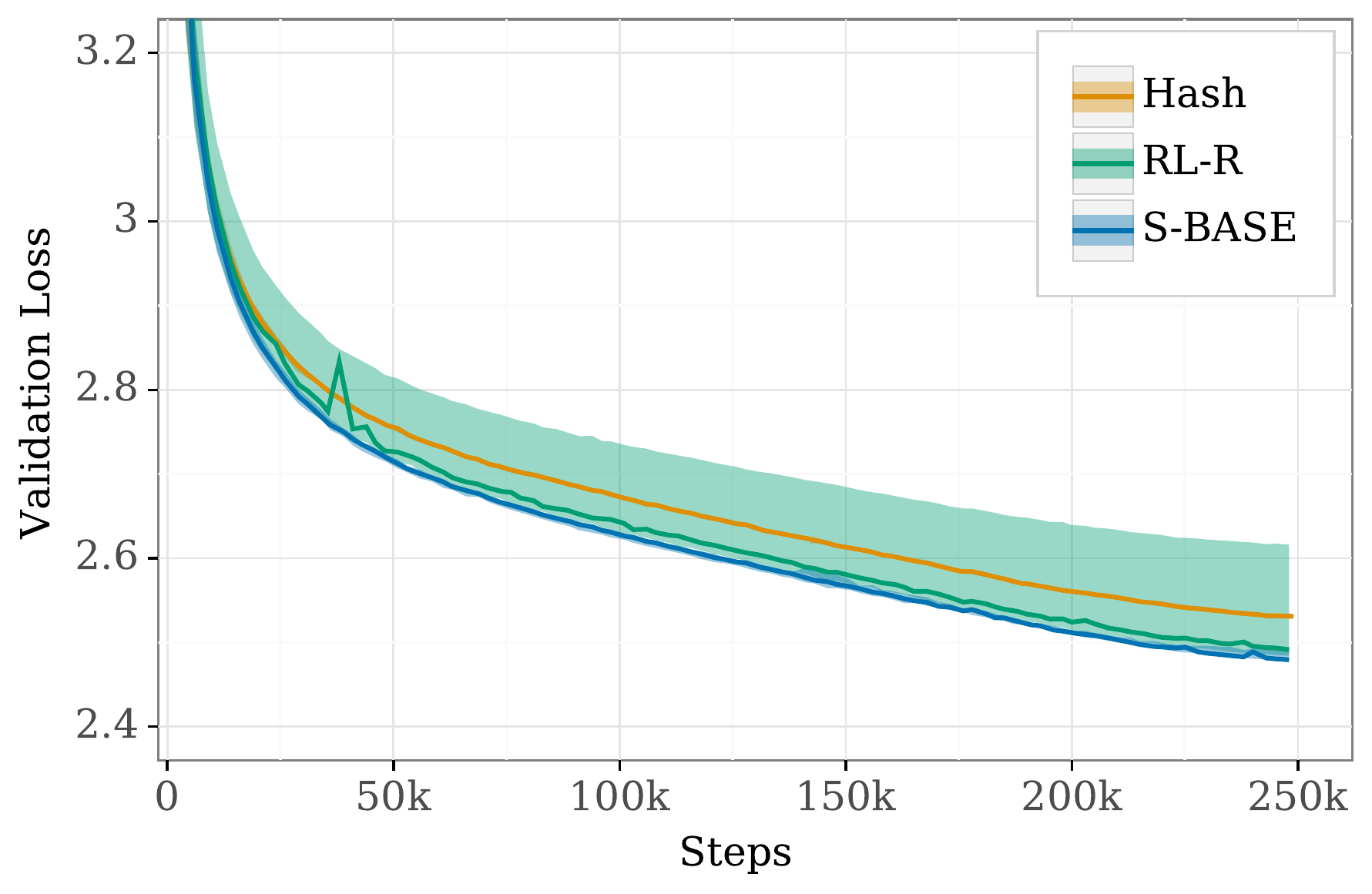}
\caption{Hyperparameter sensitivity at 512E 55M. For \rlr, hyperparameter selection has the largest impact on model performance of the three methods. The top performing \rlr models outperform \hash and are comparable with \base. However, non-optimal \rlr configurations perform worse than the other two methods.}
\label{fig:robustness}
\end{figure*}

We evaluated the robustness of \base and \rlr to changes in hyperparameters in \autoref{fig:robustness}. We focus on $E = 512$ due to anecdotal experience that the largest performance variance occurred at this scale. \rlr is found to be highly sensitive to the hyperparameters in Table \ref{tab:hyperparameters}, especially the choice of balancing weight. In addition, changes to the policy entropy weight can lead to unbalanced routers when the balancing weight is not tuned jointly.

Unlike Switch which has been shown to be sensitive to the choice of balancing loss \citep{roller2021hash}, \base is robust to changes in balancing weight for values of $1e-3$ to 1. \base also has competitive performance without a balancing loss, but training is less stable. Additionally, Switch has higher expert oversubscription rates even when tuning the balancing weight. 

\subsection{Varying Routing Frequencies}
\begin{figure*}[t]
  \centering
  \includegraphics[ width=0.8\textwidth]{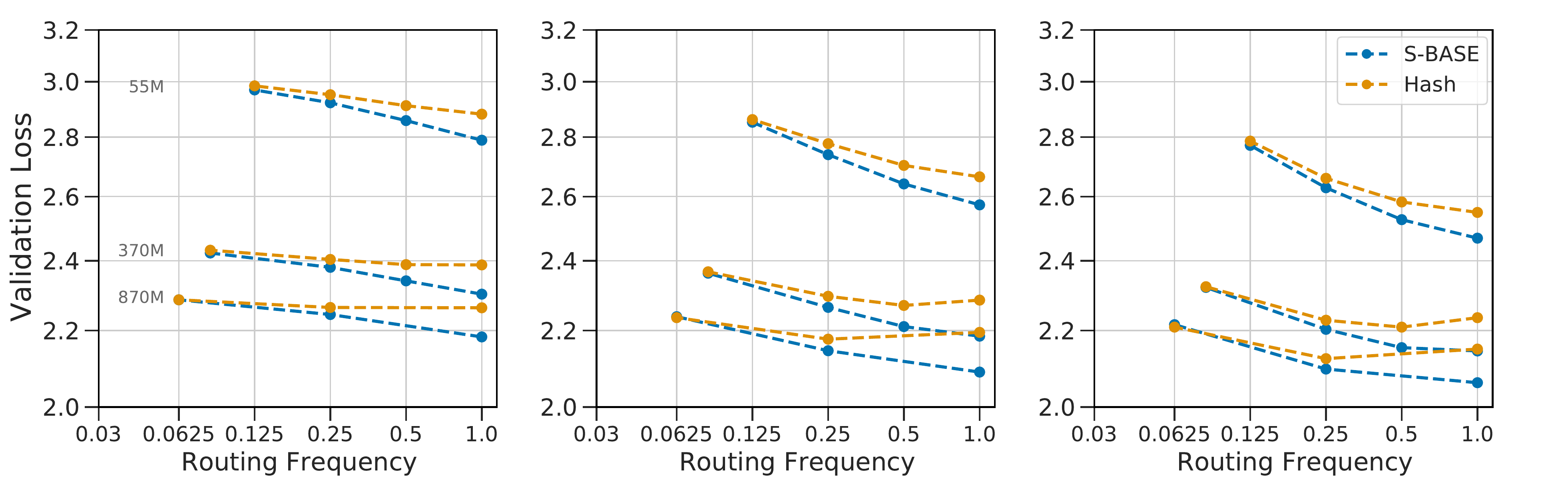}
\caption{The model performance improves with increasing routing frequency for \base, while \hash flattens at higher frequencies for 8E (left), 64E (middle) and 256E (right).}
\label{fig:routing_freq}
\end{figure*}

\begin{figure*}[t]
  \centering
  \includegraphics[width=0.4\textwidth]{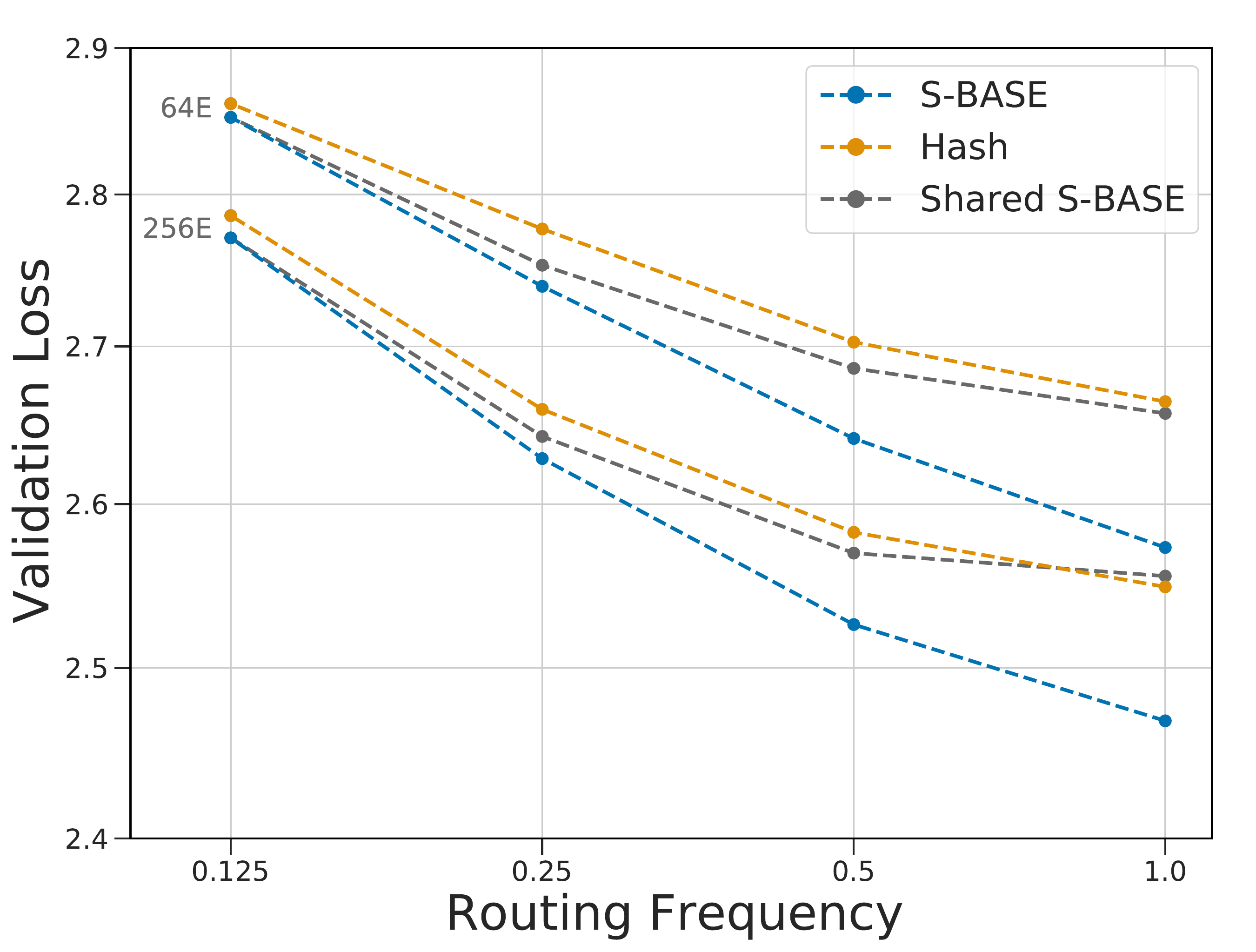}
\caption{Shared expert selections across layers has a large effect on performance for \base (in grey) at 25M. \base scales similarly to \hash in the single router case.}
\label{fig:single_expert_freq}
\end{figure*}

All of our models thus far have been routed every other layer with experts which are single FFWs \citep{lepikhin2020gshard, fedus2021switch}. However, \citet{lewis2021base, roller2021hash} explored stacking FFWs in the experts and placing $N$ routed layers at $\frac{L}{N+1}...\frac{NL}{N+1}$. We consider the performance impact of alternative routing frequencies, varying the frequency $R=\frac{N}{L}$ and placing routed layers at $\frac{L}{N}...\frac{NL}{N}$. 

We compare routing every layer to routing at frequencies $R \in \{\frac{1}{2}, \frac{1}{4}, \frac{1}{L}\}$. For routing a single layer we chose the second to last layer \citep{roller2021hash}, but consider routing at $\frac{L}{2}$ in subsection \ref{sssec:route_one_layer}. \base scales well with routing frequency, but \hash degrades in performance as shown in \autoref{fig:routing_freq}. At a single routed layer, \hash has the lowest validation loss across model sizes.

\subsection{Varying the Routing Policy}
Motivated by the improved scaling results for \base, we investigate whether learning a routing policy becomes more beneficial as the frequency of routers increases. 

\paragraph{Shared routing decisions.} In \autoref{fig:single_expert_freq}, the routing decisions are made at the first routed layer and shared across layers, which keeps the number of routers constant as $R$ increases. As \hash selects experts based on the token index at the input layer, its routing function is unchanged for this variant. \base and \hash have similar losses for shared routing decisions, whereas \base improves when learning to route at each expert layer.

\paragraph{Permuting the hash function.} Conversely, we tested a variant of \hash where the hash function at each router uses a static permutation of the input tokens to select the experts. This allows tokens to be routed to the same expert at some layers without having the same hash. We found that performance was unchanged for this variant, suggesting that increasing the number of possible routing paths does not necessarily impact performance for static policies.

These router variants suggest that methods which can adapt to each expert layer will outperform static policies. Further work is needed in analyzing how policies can more effectively learn to route across layers.

\subsection{Routing a Single Layer} \label{sssec:route_one_layer}

\begin{figure*}[t]
  \centering
 \subfigure[]{\includegraphics[width=0.35\textwidth]{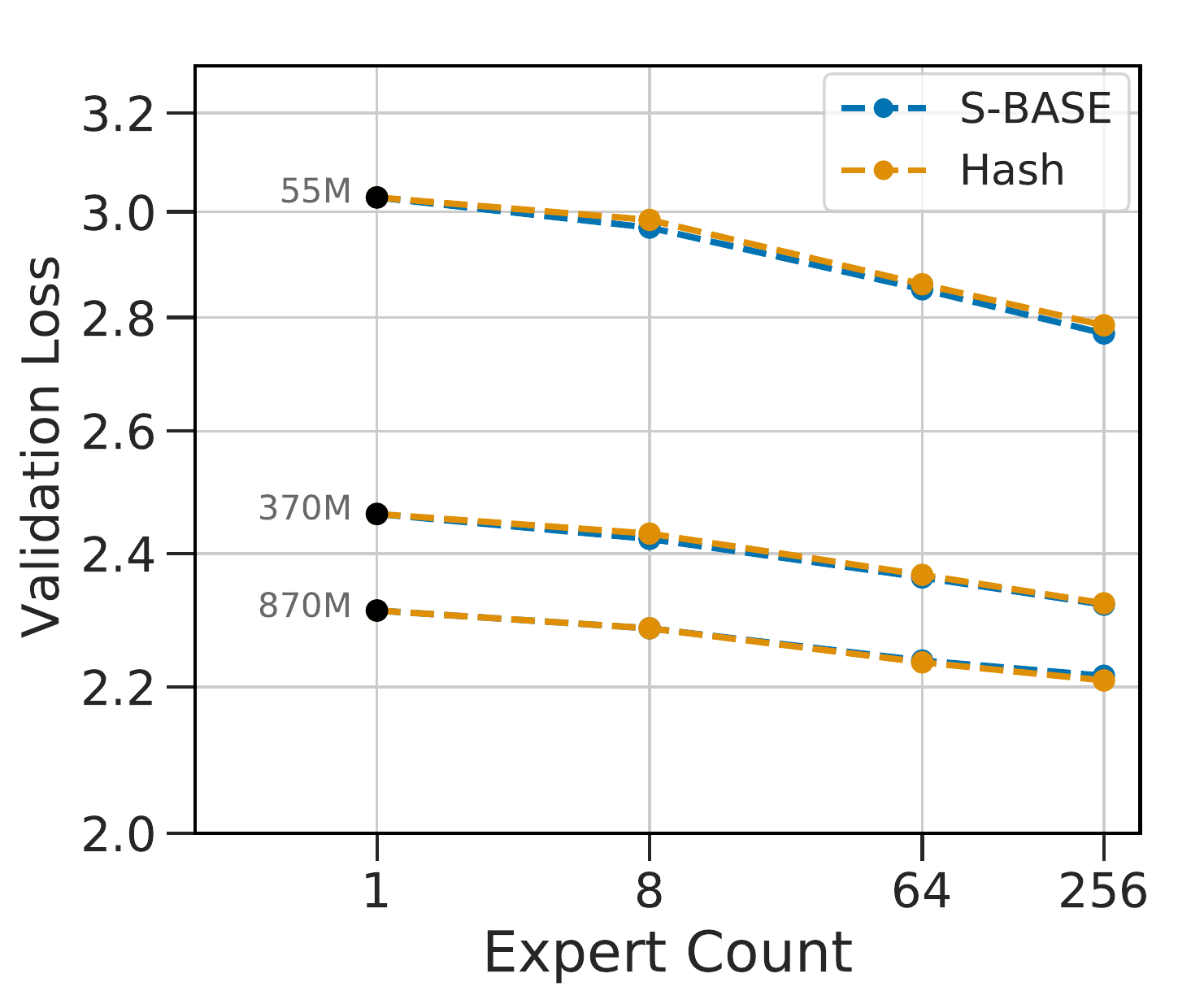}}
  \subfigure[]{\includegraphics[width=0.35\textwidth]{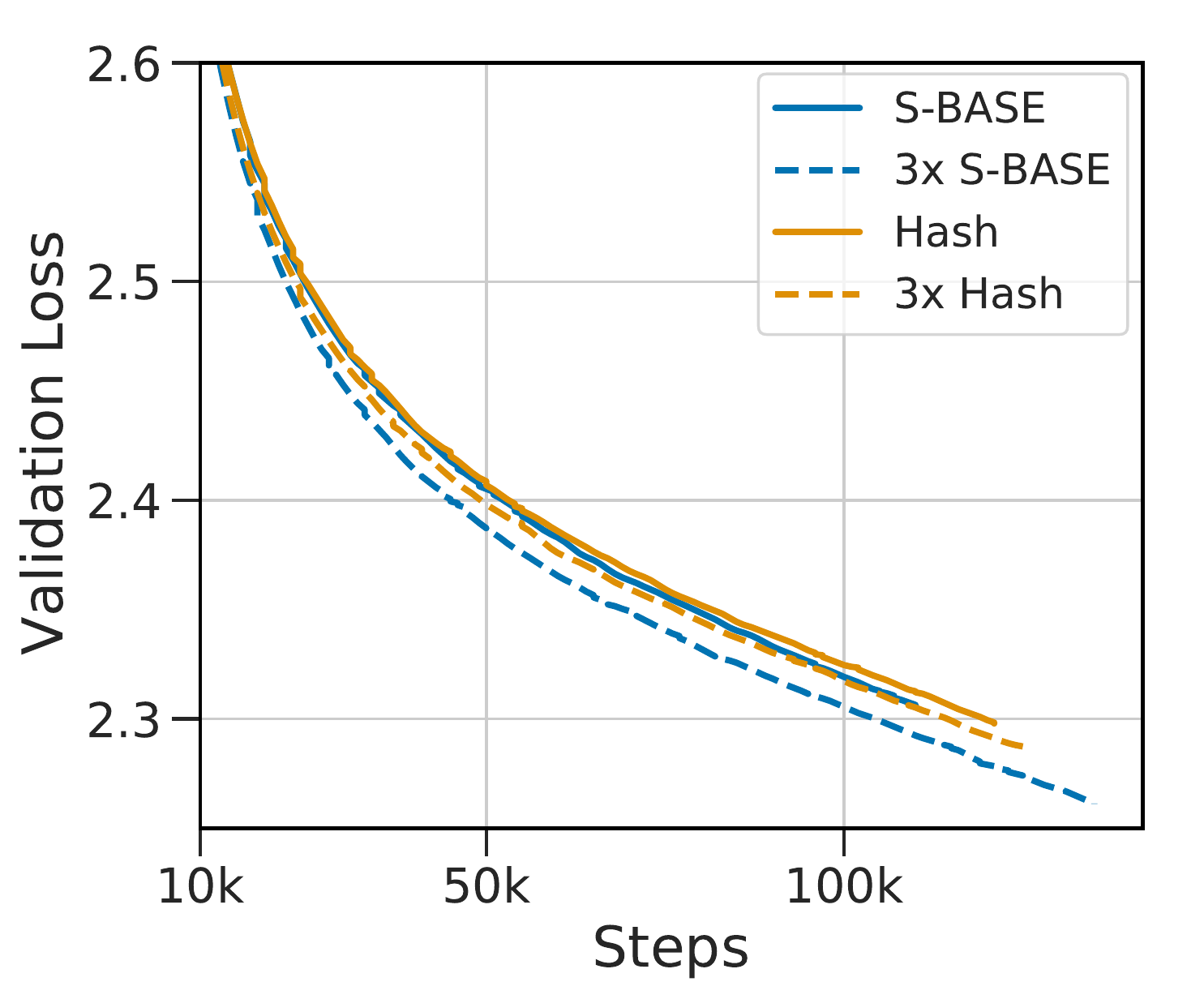}}
\caption{\textbf{(a)} \base and \hash scale similarly when routing a single layer at $L-1$. \textbf{(b)} We see similar performance for \base and \hash at 32E 1.3B when routing at $\frac{L}{2}$ with three FFWs per expert. However, \base performance is improved for interleaving three routed layers.}
\label{fig:num_expert_layers}
\end{figure*}

We analyzed the scaling behavior of \hash and \base when only routing a single layer. We observed that the routing gains for $R=\frac{1}{L}$ deviated from higher frequencies, which also impacted $R=\frac{1}{4}$ to a lesser degree. We attribute this performance regression to the suboptimal behavior of the first routed layer. In both cases the total number of routers is low, and the first layer has a larger impact on overall performance than at higher routing frequencies. For $R=\frac{1}{L}$ the complexity of routing is reduced and a simpler routing method can reach competitive performance. \hash and \base have similar performance across expert counts in this case, as shown in \autoref{fig:num_expert_layers}. 

We also compared routing a single layer at $\frac{L}{2}$ with three FFWs per expert to three evenly spaced routed layers in \autoref{fig:num_expert_layers}. Similar to the results shown in \citep{roller2021hash}, three evenly spaced routed layers has slightly better performance than three stacked FFWs for a 32E 1.3B model. We also found that \base benefits more from interleaving the routed and dense layers, which is consistent with our routing frequency results.
 
\subsection{Varying number of experts per datapoint}
 
\begin{figure*}[t]
\centering
\includegraphics[width=.8\textwidth]{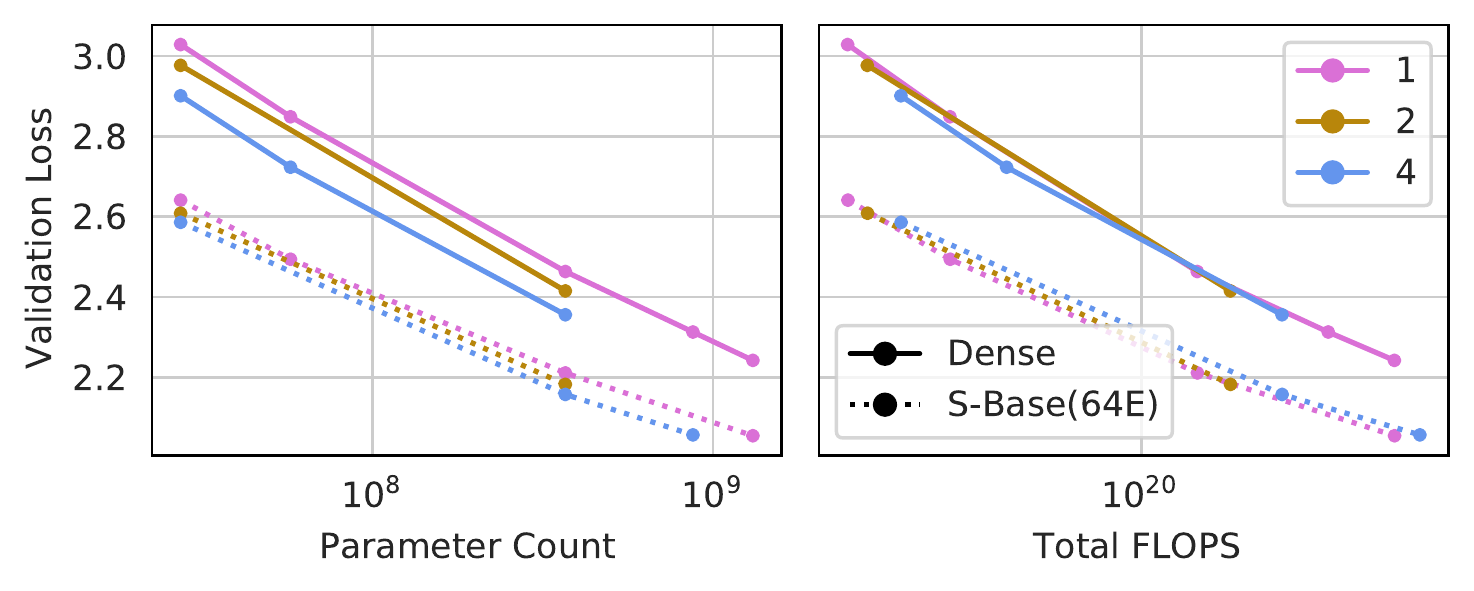}
\caption{
Example of scaling curves for a dense model and a \base(64E) model. When looking at performance per parameter, higher values of K are always better. But lower values of K are generally more flop efficient, and achieve a better loss for a given FLOP budget. 
}
\label{fig:scaling-per-k}
\end{figure*}

 In this work we have focused on routing each datapoint to a single expert at all routing layers, \emph{i.e.} for the case where $K=1$. However, SMoE models have historically routed datapoints to more than one expert \citep{shazeer2017outrageously,lepikhin2020gshard,ramachandran2018diversity}. 
 Increasing $K$ incurs in extra computation on the experts, but this additional computation may be helpful for the end result, reflecting in better loss.
 Moreover, routing a datapoint through more experts means each expert gets to see more data for each forward pass, which may speed up training. For these reasons, it is not obvious that $K=1$ is the best setup. Section \ref{sec:generalizations} investigated this and argued both that the generalized formula Equation~\eqref{eq:real_joint_scaling_law_fp} can accommodate such cases and also that the resulting fits show no substantial difference in performance for $K$. However we explore this variance more in \autoref{fig:scaling-per-k}: plotting both scaling curves for varying values of $K$ as well as plotting the loss in terms of $F$. Higher values of $K$ invariably yield better performance per step, but they are not necessarily more flop efficient. In fact, $K=1$ is always in the pareto front. We can verify that this holds for varying numbers of experts. 
 
 Note that this difference in flop-efficiency is not only theoretical, and is also followed by increased communication costs when using expert parallelism. We observed in practice that reducing K by half amounted to close to 2x speedup in inference and training.

\section{Effects of scaling strategy on Zero-shot Transfer}
\label{apsec:transfer}

There is a strong relationship between the validation loss we have been discussing and the downstream performance of models and specific tasks \citep{kaplan2020scaling}.
However, recent work has shown that this relationship is not as straightforward for large Routing Networks, and individual tasks can benefit more or less from expert scaling. For example, \citet{artetxe2021efficient} show a narrowing performance gap between a \smoe Routing Network with $E=512$ and its dense equivalent, with more marked improvement from routing in some tasks like \textit{HellaSwag} and \textit{PIQA} than in in tasks like \textit{Winogrande} and \textit{ReCoRD}. Likewise,
\citet{fedus2021switch} shows that Switch benefits more from scale better in \textit{TrivaQA} than in \textit{SuperGlue}.

A detailed analysis of the scaling properties of Routing Networks and how that transfers to downstream tasks merits dedicated work. Here we start the conversation by looking at zero-shot transfer on a set of well known downstream tasks: \lambada~\citep{paperno2016lambada}, \pile~\citep{gao2020pile}, \cc~\citep{curation2020cc}, \wikitext~\citep{merity2016pointer} and \cfour~\citep{raffel2020exploring}. 

We estimate the scaling coefficients individually for each task and routing technique. For simplicity of interpretation we ignore the bounded scaling term and focus on the bilinear fit on Eq.~\ref{eq:no_tailoff}. The coefficients can be seen in Table~\ref{tab:transfer_coefficients}. We expect that scaling in both $N$ and $E$ will improve the downstream performance. The key question revolves around understanding changes in the relative magnitude of $a$, $b$ and $c$ as we move from task to task.

\begin{figure*}[th]
  \centering
  \includegraphics[width=1\textwidth]{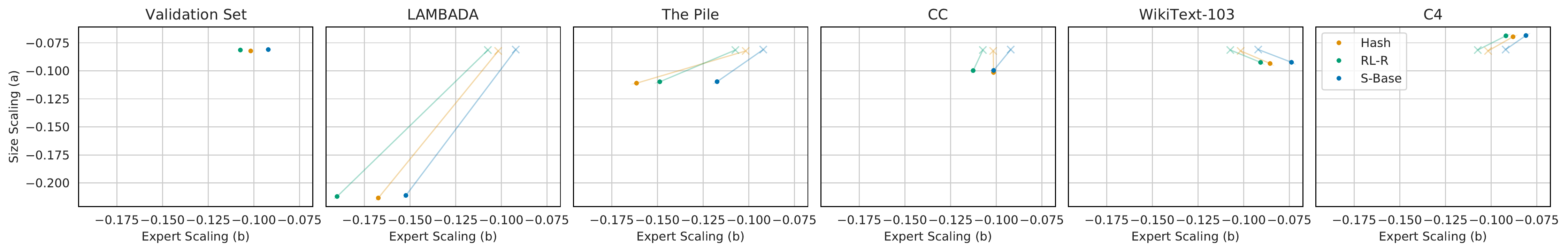}
  \caption{
  Individual scaling coefficients for different tasks and routing techniques, compared to the coefficients estimated in the validtion set. Different techniques also scale differently depending on the task, but this also depends on the interaction term (see \autoref{fig:transfer-alpha}) 
  }
\label{fig:transfer-individual-scaling-coeffs}
\end{figure*}

Viewing Table~\ref{tab:transfer_coefficients} it is immediately clear that the individual scaling coefficients vary greatly across tasks, \emph{i.e.} different tasks have different relative gains at Zero-Shot performance as we move to larger scales. This can be better shown in \autoref{fig:transfer-individual-scaling-coeffs}, where all coefficients are displayed in a single plot. The variation across tasks are not the same for $a$ and $b$. e.g. \wikitext has higher values for $b$ and lower for $a$ when compared to the validation set. This means that even though tasks see monotonic improvement in performance by scaling through either adding more experts or increasing the base model size, some tasks benefit more and some less from which method is used. 

\begin{figure*}[t]
  \centering
\includegraphics[width=\textwidth]{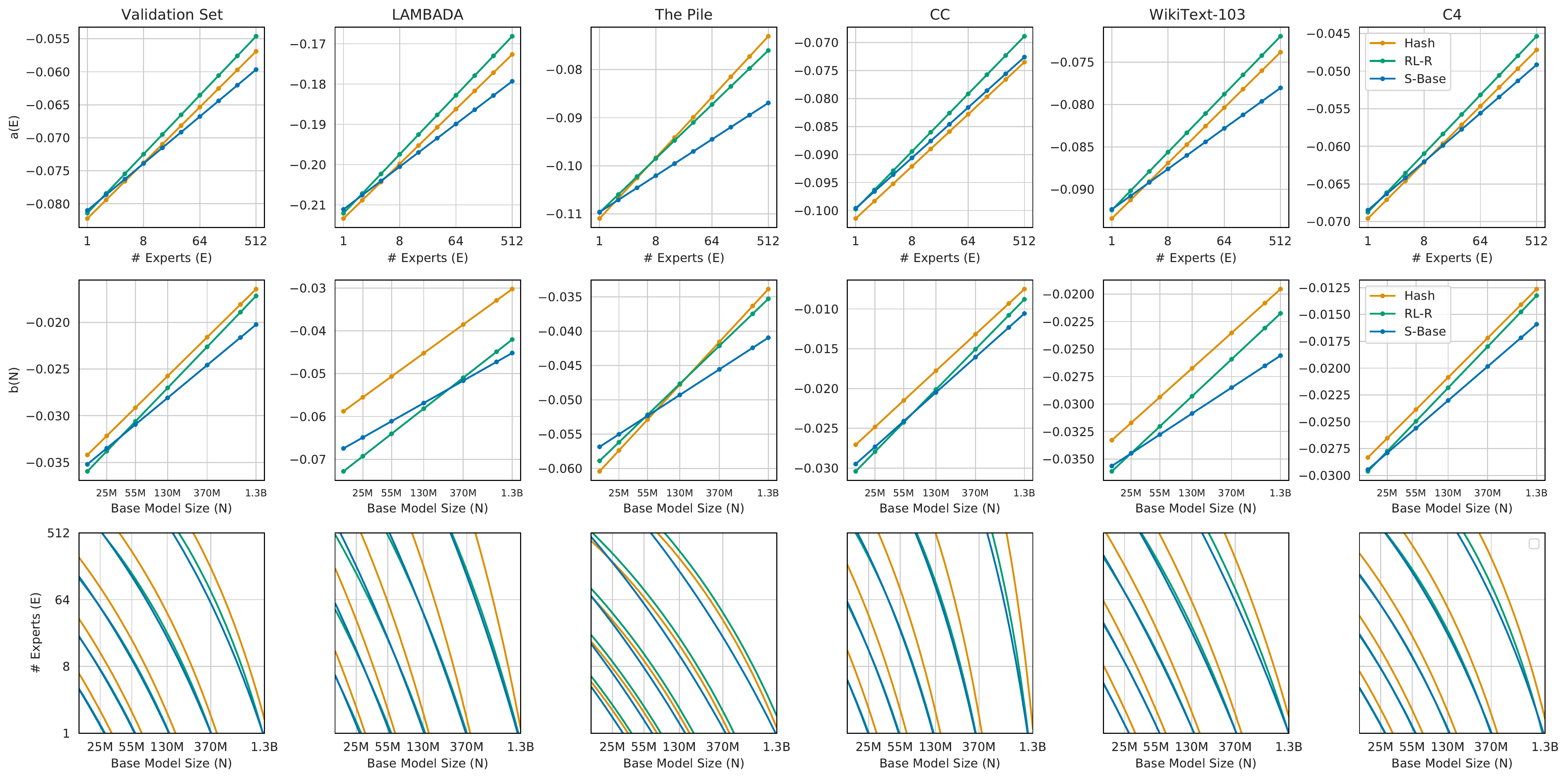}
\caption{Estimated scaling coefficent for Zero-shot performance across different datasets. Top half: The coefficient for increasing $E$ while keeping $N$ fixed for varying values of $N$ at different downstream tasks. Middle: The coefficient for increasing $N$ varying values of $E$. 
}
\label{fig:transfer-alpha}
\end{figure*}

For a more complete picture, we can account for the $N$ and $E$ interaction coefficient $c$ by incorporating it into one of the scaling coefficients -- by holding the other quantity fixed -- which leads to $a(E)$ and $b(N)$ (see Section~\ref{subsec:quadratic}). This can be seen in \autoref{fig:transfer-alpha} for varying values of $N$ and $E$.

We see that \base tends to dominate with lower coefficients at higher values of $E$ and $N$ (due to its smaller interaction term relative to scaling terms), but this varies across tasks. For example, \rlr shows better $b(N)$ for most values of $N$ in LAMBADA, until it is overtaken by \base at $N=410M$, but \base is always superior in C4.
Moreover, the ordering is not consistent between \hash and \rlr across tasks, even though they often do not cross.
This all means it is difficult to establish superior performance of a routing technique without looking at a variety of tasks and scales.

We often want to compare Routing Networks with a dense baseline with the same performance on the validation set, and see how this changes on downstream tasks.
We can use these parameters in a simplified version of the Effective Parameter Count (EPC, Equation~\ref{eq:equivalence_tail_off}), by assuming $E_{start}=1$ and $E_{max}=\infty$, such that $\hat{E}=E$.
First, we note that since the coefficients vary greatly across tasks, each task will have a different EPC for the \emph{same network configuration}.
Moreover, the effects of scaling by varying $E$ and $N$ will vary across tasks.
Say we have a routing net of size $N$ with $E$ experts and we want to increase its base model size by a factor of $x$ while keeping the same number of experts. The effect on $\bar{N}$ in this case will be a multiplication by $x^{a(E)/a(1)} = x^{1 + c \log E}$. Since $c$ varies per task, the improvement achieved by increasing the base model size will also be task dependent.  

Say we have a routing net of size $N$ with $E$ experts and we want to increase its base model size by a factor of $x$ while keeping the same number of experts. The effect on $\bar{N}$ in this case will be a multiplication by $x^{a(E)/a(1)} = x^{1 + \frac{c}{a} \log E}$. Since $\frac{c}{a}$ varies per task, the improvement achieved by increasing the base model size will also be task dependent.

For example, the \epc$_{validation}$ for N=110M, E=32 is 370M, but \epc$_{lambada}$ for the same model is 284M, while \epc$_{pile}$ is 535M. The key implication here is not only do the values change, but their slopes are different. This means that downstream tasks must be analyzed carefully: a practitioner could scale a model via routing expecting some overarching improvement, but get a much diminished (or enhanced!) improvement on specific downstream tasks, depending on their specific values of $b$ and $c$.

\section{On Convergence, or Lack Thereof}
\label{apsec:convergence}

Here we digress on two important details, both focusing on token count.
First we argue that discussing converged performance of large transformers on modern and massive text datasets is probably a misnomer; scaling analyses should focus on optimal performance at a fixed number of tokens. Second, we provide evidence arguing against a proposed equation in \citet{kaplan2020scaling} (Eq.~(1.6)).

\subsection{Convergence on Large Datasets}
\label{apssec:large_scale_convergence}

There are two cases where the converged performance of a model can be clearly defined. The first is when continued training of the model produces no improved results (even analyzed at logarithmic scale in the number of tokens), the second is when continued training leads to reduced validation performance: overfitting.

Our models exhibit neither behavior. No overfitting is seen even for our largest models, likely due to the complexity and size of the dataset used. Furthermore, despite being trained for 130 billion tokens, not even our smallest models have saturated. We push this envelope even further:
training two additional sets of \texttt{15M} models with $1$, $64$, $128$ and $512$ experts. The first set is trained for just $75,000$ steps, and the second for $1,000,000$ steps: four times more data (half a trillion tokens). We highlight that this involves corresponding changes to the cosine cycle decay. We exclusively train \hash models, both due to limits in the number of extra models we were able to train and also because it has the largest value of $\Emax$.

\begin{figure*}[t!]
  \centering
  \includegraphics[width=0.8\linewidth]{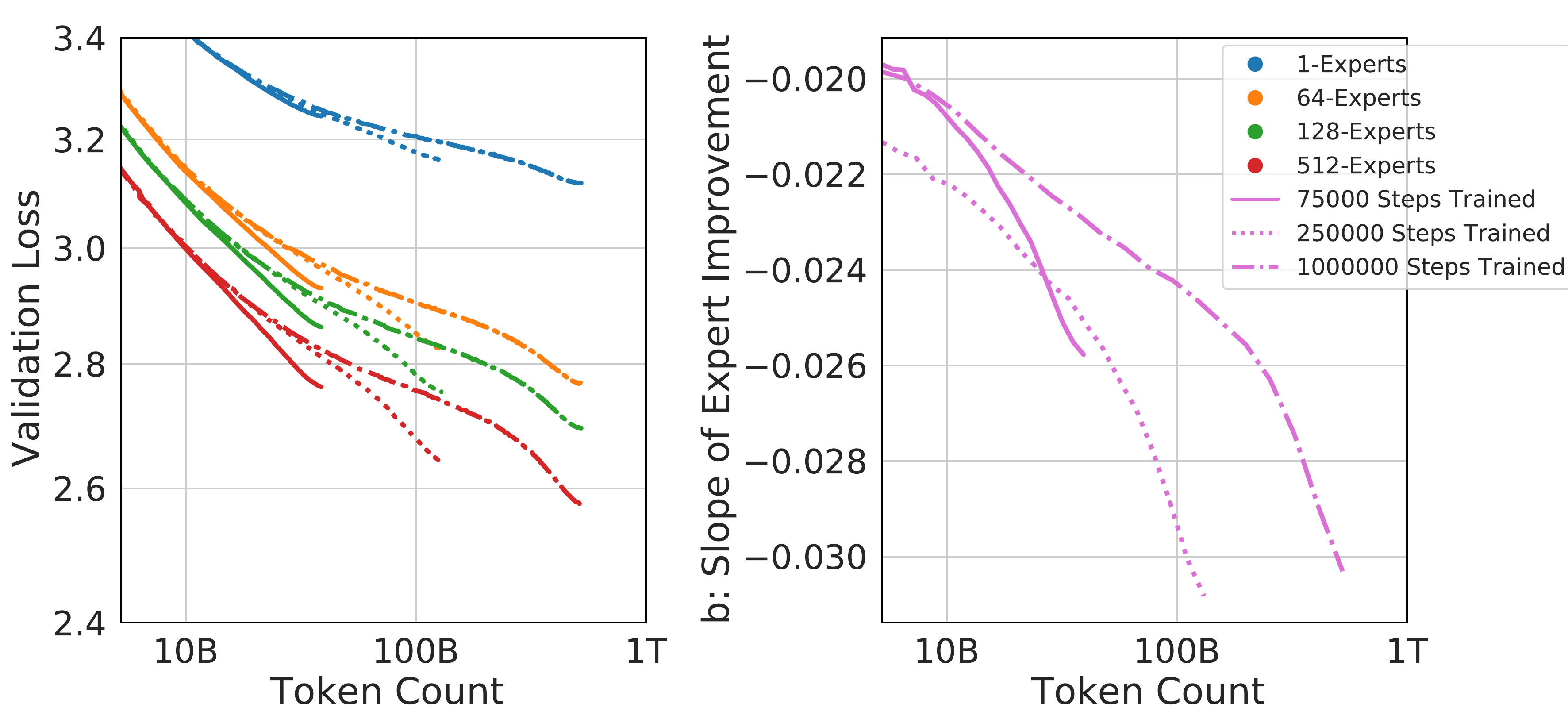}
\caption{On the left: validation performance over time for $15M$ models trained with different expert counts and over three different lengths of training. On the right, the coefficient $b$ from fitting Eq.~\eqref{eq:hopeful_power}, representing scaling from $E$ across intermediate values.}
\label{fig:different_lengths}
\end{figure*}

Results from these models are plotted in \autoref{fig:different_lengths} (left). \texttt{15M} with no routing, the smallest model we train as part of this work, is still far from having saturated its performance capability. Indeed, training for 4x longer further reduces the validation loss by $0.05$. This pattern continues, and is exacerbated, when increasing the expert count: the same model with $E = 512$ gets a $0.07$ reduction in loss from 4x more tokens. 

It is clear then that the very smallest models considered have yet to converge. The same is certainly true for larger ones, and probably more so. If 500 billion tokens is a lower bound to the convergence point of \texttt{15M}, the analysis in \citet{kaplan2020scaling} would predict needing trillions of tokens to converge \texttt{1.3B}: much more than what was used to train some of the largest language models yet created \citep{brown2020language}. For large, complex text datasets of the scale used to train large language models, convergence is not a proper criteria.

\subsection{Performance Qualified on Token Count}

Rather than claiming analysis at a non-observed point of convergence, we emphasize that the scaling behavior we have described in this work is valid only as a function of a particular number of steps (or tokens). At each point, we can define instantaneous values of scaling coefficients, with the values from all models taken at $S$ steps\footnote{This sidesteps the issue of critical batch size \citep{mccandlish2018empirical, kaplan2020scaling}, consideration of which requires a substantially larger sweep of models. Future work estimating the critical batch size will likely lead to better model fits.
}.

In fact, the situation is more complicated that simply conditioning our scaling coefficients on token count. We can see this by plotting $b$, the scaling coefficient for changes in expert-count in \autoref{fig:different_lengths}(right). An immediate observation is that the values of $b$ are non-constant, supporting the need to qualify scaling on token count. A second, more substantial point, is that these values are not uniquely defined by token count. For a given number of tokens, the scaling behavior of three different sets of models is completely different, dependent on how far into the learning rate schedule those sets of models were. We note that this behavior is suggested by experiments in \citet{kaplan2020scaling} (App. D.6).

Attempting to find the full set of parameters on which these scaling terms depend is beyond the scope of this work. We highlight just the importance of insuring that all variables possible are matched when comparing values to calculate scaling coefficients.

\subsection{Performance Modeled as L(N, S)}

We conclude by highlighting one implication of the fact that scaling coefficients are dependent on token count. We analyze only the dense models trained as part of this work, and calculate values of $a$ in Equation~\eqref{eq:dense_law} for all dense models trained as part of the primary sweep across all step counts; plotted in \autoref{fig:bad_fits}(a) with RMSE values plotted in \autoref{fig:bad_fits}(b). First, it is important to emphasize that the fits remain good throughout $S$ (after an initial period of transience). Namely, though the slope is different, the validation losses for a given intermediate $S$ follow a power law about as well as they do later in training (if anything, more so). Second, the estimated coefficients $a$ are clearly monotonically increasing with $S$.

\begin{figure*}[t!]
  \centering
  \subfigure[]{\includegraphics[width=0.3\linewidth, height=0.30\linewidth]{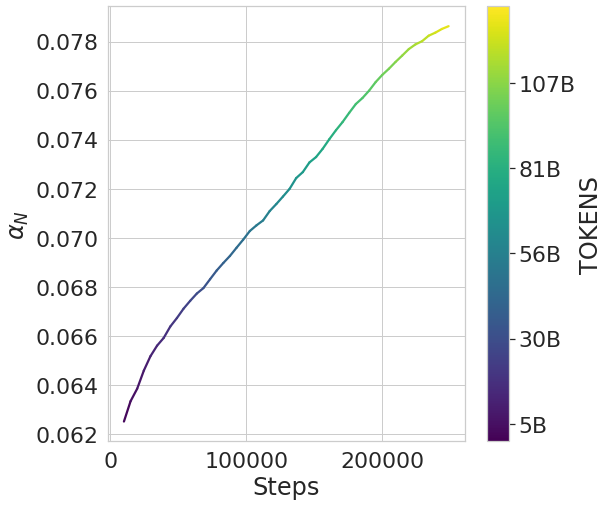}}\quad
  \subfigure[]{\includegraphics[width=0.3\linewidth, height=0.30\linewidth]{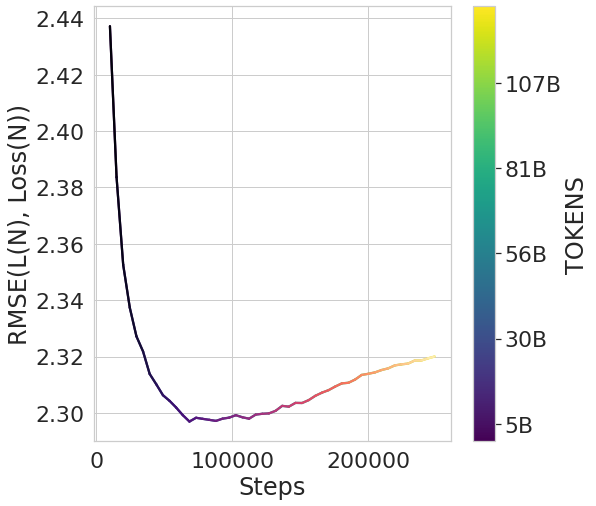}}\quad
  \subfigure[]{\includegraphics[width=0.3\linewidth, height=0.30\linewidth]{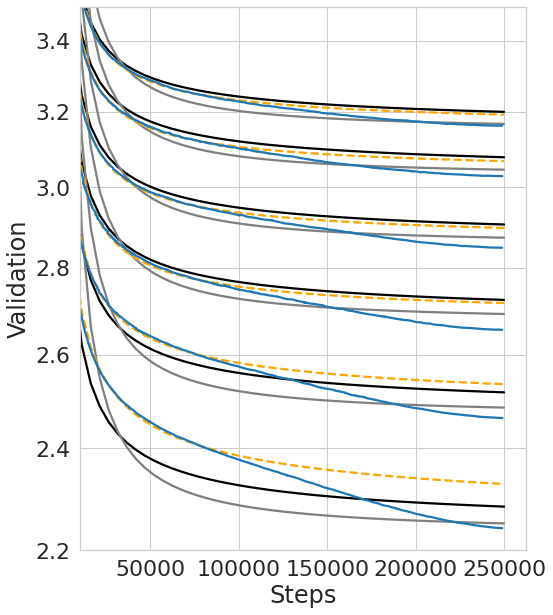}}
\caption{
\textbf{a)} Values of $a$ found for dense models across training.
\textbf{b)} RMSE for these same fits.
\textbf{c)} Three attempts to fit Eq.~\eqref{eq:bad_l_n_s}. In black the standard fit. In orange and grey fits only using and ignoring the final $150,000$ steps respectively.
}
\end{figure*}
\label{fig:bad_fits}

\citep{kaplan2020scaling} propose (Eq. 1.6) a unified prediction of the loss achieved by a model with size $N$ training for $S$ steps:
\begin{equation}
    L(N, S) = \left(\frac{N_c}{N}\right)^{\alpha_N} + \left(\frac{S_c}{S}\right)^{\alpha_S}
\end{equation}
\label{eq:claimed_step}
This comes with the subtlety that $S$ must be defined as the number of steps \textit{when training at the averaged critical batch size}, where our models are trained with a fixed batch size. This means a proper analysis must use $S_\mathit{min}$ with $S = S(1 + \frac{B_c}{B L^{-\alpha_B}})^{-1}$ for constants $B_c$ and $\alpha_B$. It is important to highlight however that $S_\mathit{min}$, as described in \citet{kaplan2020scaling}, should be independent of $N$. This implies that $\frac{\partial}{\partial N}\left(L(N, S)\right)$ is independent of S, or in log-log space:
\begin{equation}
    \left. \frac{\partial \log(L(N^*, S))}{\partial N^*} \right|_{N^* = 10^{N}} = -\alpha_N\label{eq:bad_l_n_s}
\end{equation}
This prediction of constant scale is in concrete opposition to the increasing value seen in \autoref{fig:bad_fits}(a). We can furthermore check that this functional form cannot be obviously fit to our learning curves, with examples show in \autoref{fig:bad_fits}(c). 

There are subtle differences between training setups, and we do not want to claim our experiments wholly disprove the conjecture in \citep{kaplan2020scaling}. However, the results in \autoref{fig:bad_fits} motivate us to assume that Eq.~\eqref{eq:claimed_step} cannot be used to model our specific training curves. A consequence of this is that we can also no longer conclude Equation B.5 from \citep{kaplan2020scaling}, that:
\begin{equation}
    L(N_{eff}(C), C) = (1 + \frac{\alpha_N}{\alpha_S})L(N_{eff}, \infty)
\end{equation}
With this equation, we might be able to lower-bound true converged performance (which we have not seen in our models) by inference from compute-efficient performance, which has been achieved by the majority of our models.

\section{Large Scale Routing Behavior, Coefficient Sensitivity, and Future Work}

\begin{figure*}[t!]
  \centering
  \includegraphics[width=0.5\linewidth]{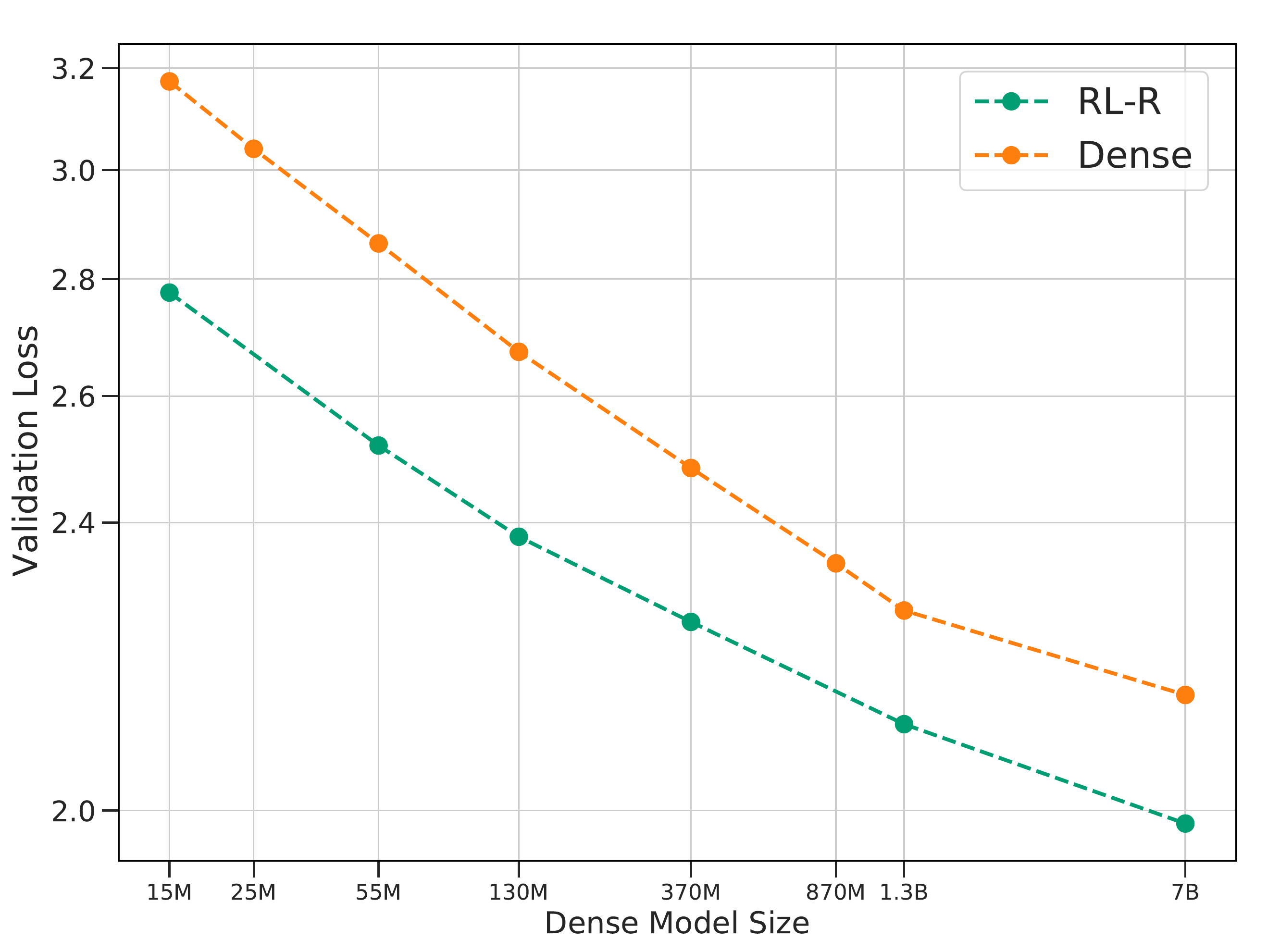}
\caption{\rlr performance for 64E continues to scale well compared to dense up to 7B base model size.} 
\label{fig:7b_scaling}
\end{figure*}

Our analysis predicts that larger values of $E$ will continue to improve performance, especially for small models, at a diminishing rate. \autoref{sec:n_max} also predicts that routing will continue to help with increasing $N$ for at least one, if not two orders of magnitude larger base model size. Practical compute limitations prevented our sweep from exploring these regimes, and there are interesting unanswered questions in the limit of these two variables. In particular, exact predictions of $\Ncut$ are highly dependent on the precise value of $b$, where error in the second decimal place shifts predicted values by orders of magnitude (not surprising, as it is the slope of a line in log-log space).

We believe exploring the limit behavior of $N$ and $E$, especially arriving at a more precise value of $b$, is crucial. Anecdotally, we can report the results of one experiment: a large \rlr model with $N > 7,000,000$, providing a rough upper bound for error in $b$ for \rlr. In particular, we trained a model with $d_{model}=4096$, $n_{layers}=32$, $n_{heads}=32, E=64$ and K/V size of 128. There are some important eccentricities of this model which affect its match to the fits described in this work: it was trained with a batch size of 1024 for 100k steps with a policy gradient weight of 1e-1 and balancing weight of 1e-1. Other training details are consistent with Section \ref{ssec:training_details}.

The performance of this model, relative to a dense model of the same size and also to a number of smaller models, is plotted in \autoref{fig:7b_scaling} evaluated at 100B tokens. The changes described above prevent the analysis in this work from accurately predicting this model's performance, but one key feature remains: the routed 7B model substantially outperforms the baseline. This is of particular interest since just a $0.01$ decrease in $b$ would predict an $\Ncut$ at $9B$, meaning we would already be close to the regime where routing would cease to work. Nevertheless, at this value routing is clearly still a major improvement, and our estimate of $b$ is unlikely to be a substantial overshoot.

While the differences between this model and those analyzed in the paper make concrete extrapolation impossible, it shows that routing techniques still maintain competitive improvements at almost an order of magnitude larger value of $N$ than analyzed and it is unlikely the scaling coefficients measured in this work substantially overestimate the routing technique's scalability. We encourage future work probing the limits of routing networks, both in $N$ and $E$, to better understand their properties and provide more accurate predictions of their scaling coefficients.

\section{Extra Plots and Tables}
\label{apsec:complete_plots}

This section contains some helpful visualizations and data which are not included in the main text.

\begin{table}[h]
\caption{Values of $b(N)$ with hold-out RMSEs in parentheses.}\label{tab:alphas}

\centering
\footnotesize
\begin{tabular}{ c | c c c c c }
\toprule
 & 15M & 25M & 130M & 370M & 1.3B \\
\midrule
\textbf{\base} & -0.035 & -0.031 & -0.029 & -0.024 & -0.019 \\
& (0.035) & (0.019) & (0.017) & (0.014) & (0.012) \\
\textbf{\rlr} & -0.033 & -0.031 & -0.027 & -0.022 & -0.016 \\
& (0.016) & (0.013) & (0.013) & (0.014) & (0.009) \\
\textbf{\hash} & -0.031 & -0.029 & -0.025 & -0.021 & -0.015 \\
& (0.039) & (0.029) & (0.023) & (0.016) & (0.011) \\
\bottomrule
\end{tabular}

% \subsection{Fits to Equation \eqref{eq:no_tailoff}.}
\end{table}
\begin{table}[h!]
\centering
\footnotesize
\begin{tabular}{ c | c c c c}
    \toprule
     & a & b & c & d \\
    \midrule
    \textbf{\base} & 0.079 & 0.088 & 0.007 & 1.072 \\
    \textbf{\rlr}  & 0.080 & 0.105 & 0.010 & 1.076 \\
    \textbf{\hash} & 0.081 & 0.097 & 0.009 & 1.086 \\
    \bottomrule
\end{tabular}
\caption{Fits to Equation \eqref{eq:no_tailoff}.}
\label{tab:abcds}
\end{table}

\begin{table}[h]
\centering
\begin{tabular}{ c | c c c c c c }
\toprule
 & \base & RL-R & HashLayers \\
\midrule
\textbf{4} & 0.077 & 0.075 & 0.077 \\
\textbf{8} & 0.073 & 0.073 & 0.075 \\
\textbf{32} & 0.070 & 0.067 & 0.069 \\
\textbf{64} & 0.066 & 0.063 & 0.066 \\
\textbf{128} &  0.064 & 0.060 & 0.063 \\
\textbf{256} & 0.058 & 0.056 & 0.059 \\
\textbf{512} & 0.060 & 0.053 & 0.056 \\
\bottomrule
\end{tabular}
\caption{Values of $a(E)$ for different values of $E$}.
\label{tab:all_alpha_ns}
\end{table}

\begin{figure}[H]
  \centering \includegraphics[width=0.33\linewidth]{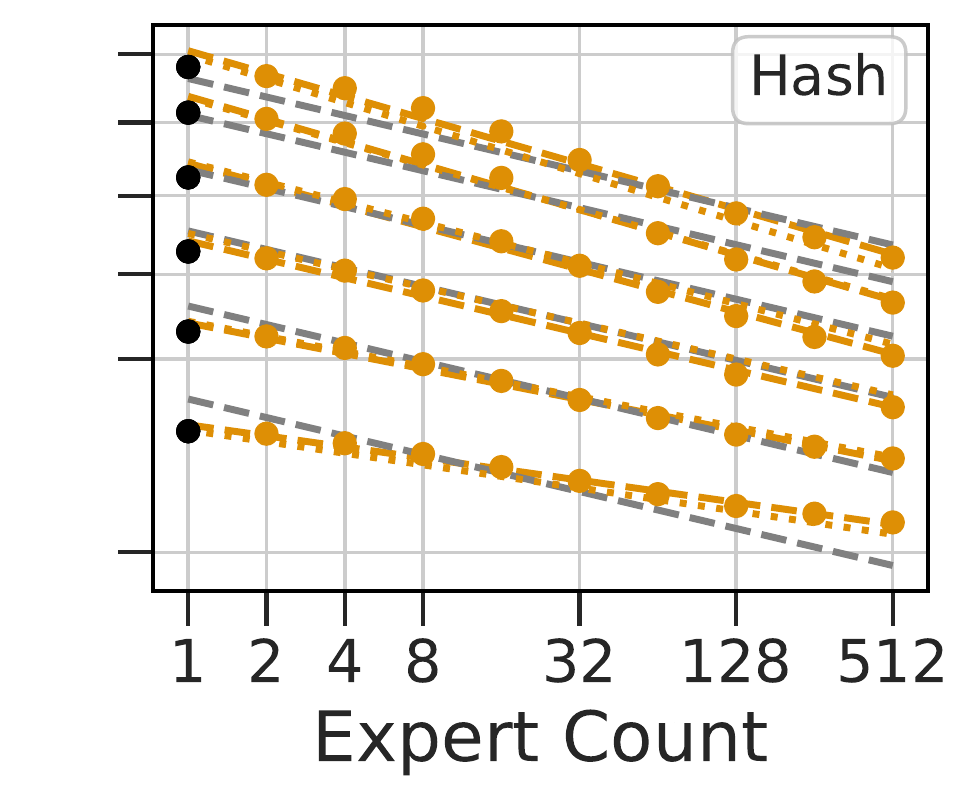}
\caption{Affine fits for \hash with a shared slope in grey.}
\label{fig:fig3_hash}
\end{figure}

\begin{figure}[h!]
  \centering
  \includegraphics[width=0.9\textwidth]{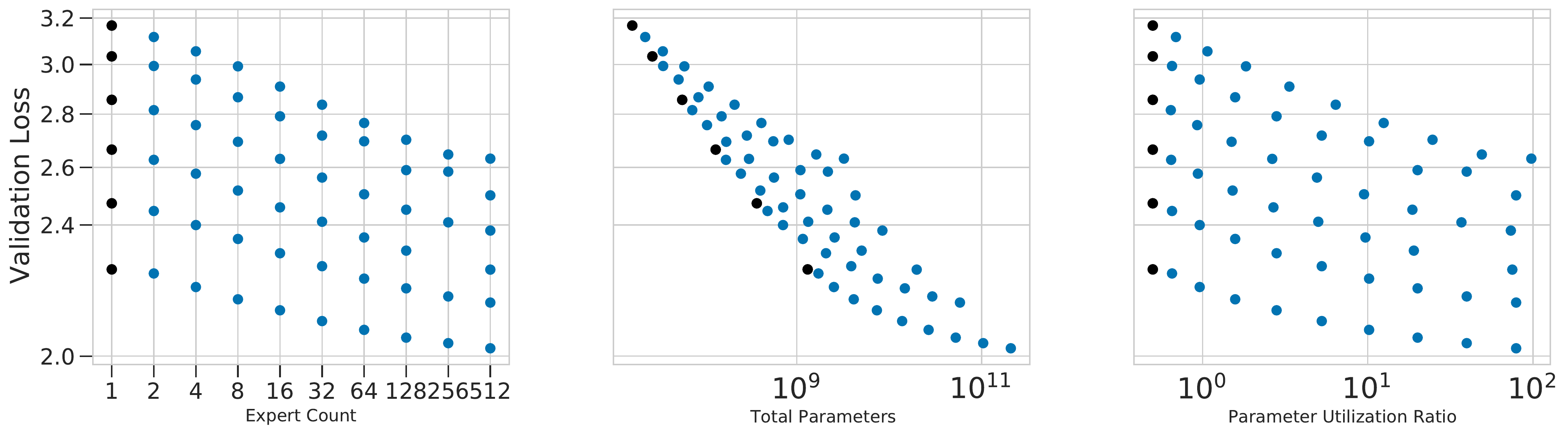}
\caption{The validation loss for all \base models plotted as a function of expert count (left), the total number of parameters (center) and the ratio of parameters to TeraFLOPs per inference (right).}\label{fig:experts_and_params1}
\end{figure}

\begin{figure}[h!]
  \centering
  \includegraphics[width=0.9\textwidth]{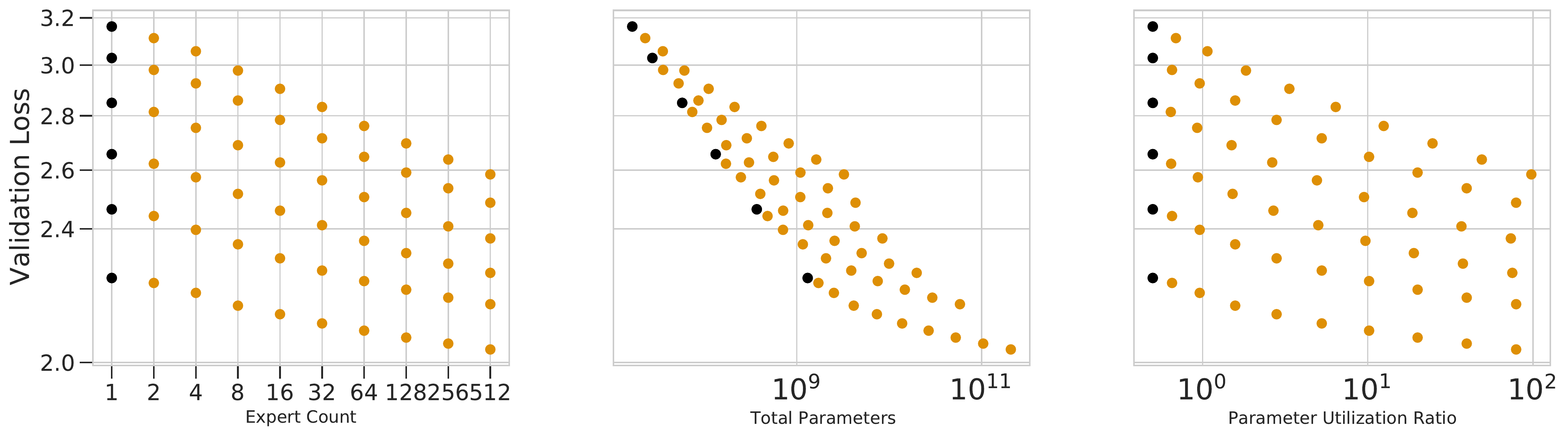}
\caption{The validation loss for all RL-R models plotted as a function of expert count (left), the total number of parameters (center) and the ratio of parameters to TeraFLOPs per inference (right).}\label{fig:experts_and_params2}
\end{figure}

\begin{figure}[h!]
  \centering
  \includegraphics[width=0.9\textwidth]{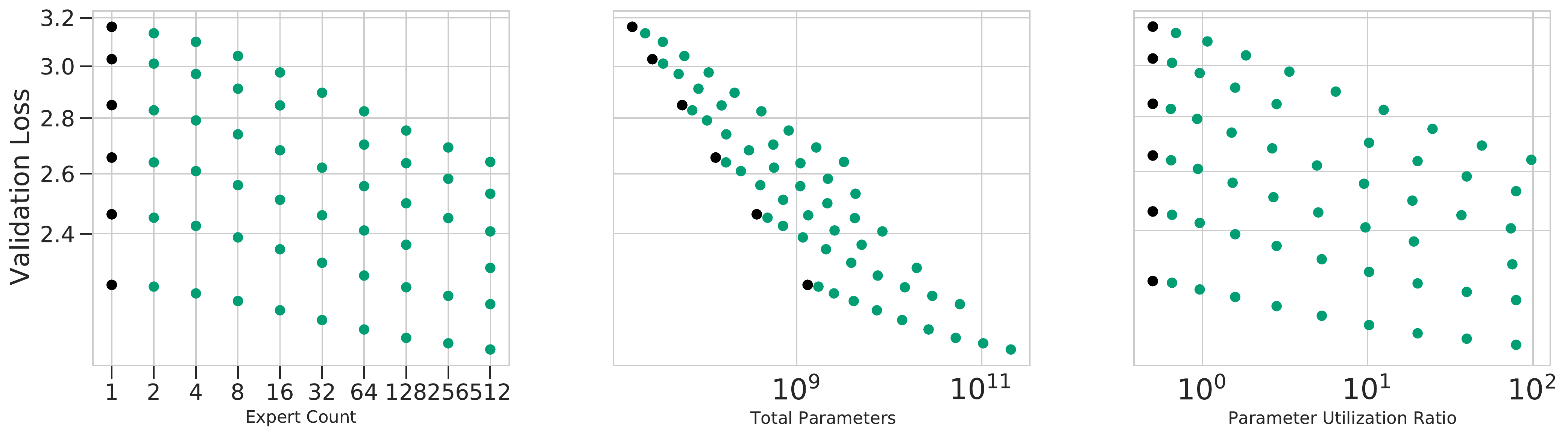}
\caption{The validation loss for all HashLayer models plotted as a function of expert count (left), the total number of parameters (center) and the ratio of parameters to TeraFLOPs per inference (right).}\label{fig:experts_and_params3}
\end{figure}

\begin{figure}[h]
  \centering
  \includegraphics[width=0.9\linewidth]{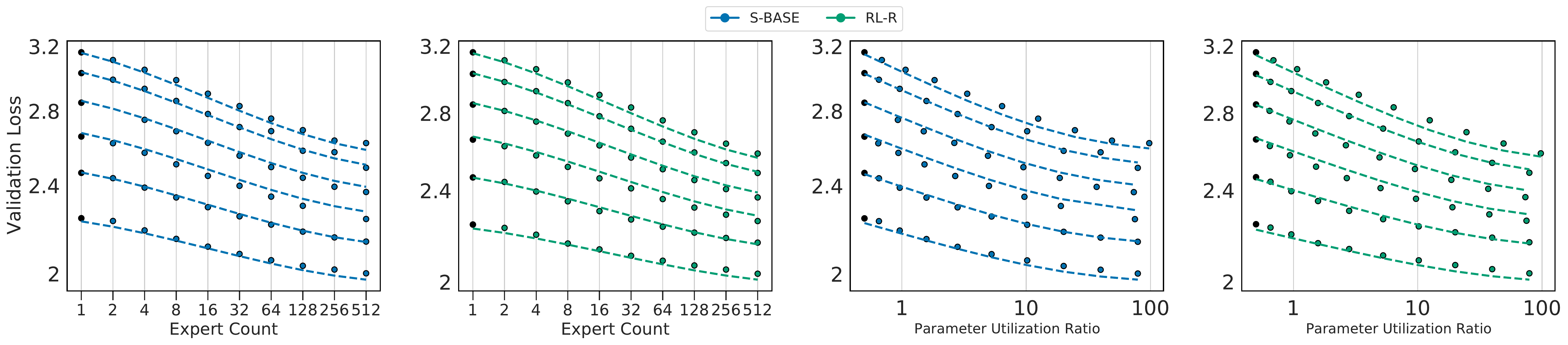}
\caption{Fitting \base and \rlr with Eq.~\eqref{eq:real_joint_scaling_law} and Eq.~\eqref{eq:real_joint_scaling_law_fp}.}
\label{apfig:pf_and_expert_fits}
\end{figure}

\begin{figure}[h]
  \centering
  \includegraphics[width=0.9\linewidth]{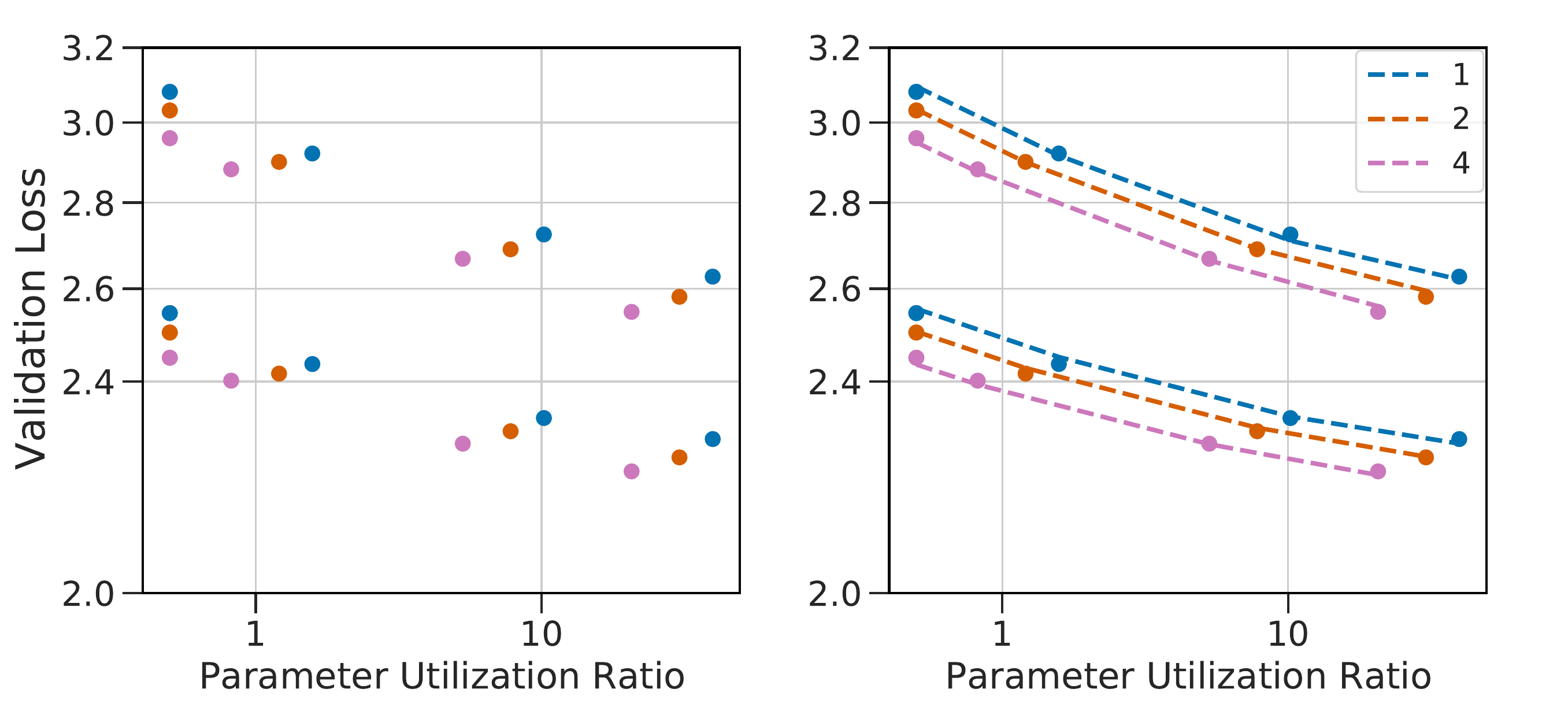}
\caption{Joint fits to Equation \eqref{eq:real_joint_scaling_law_fp} for $K \in \{1, 2, 4\}$.}
\label{apfig:pf_fits_topk_individual}
\end{figure}

\begin{figure}[h]
  \centering
  \includegraphics[width=0.9\linewidth]{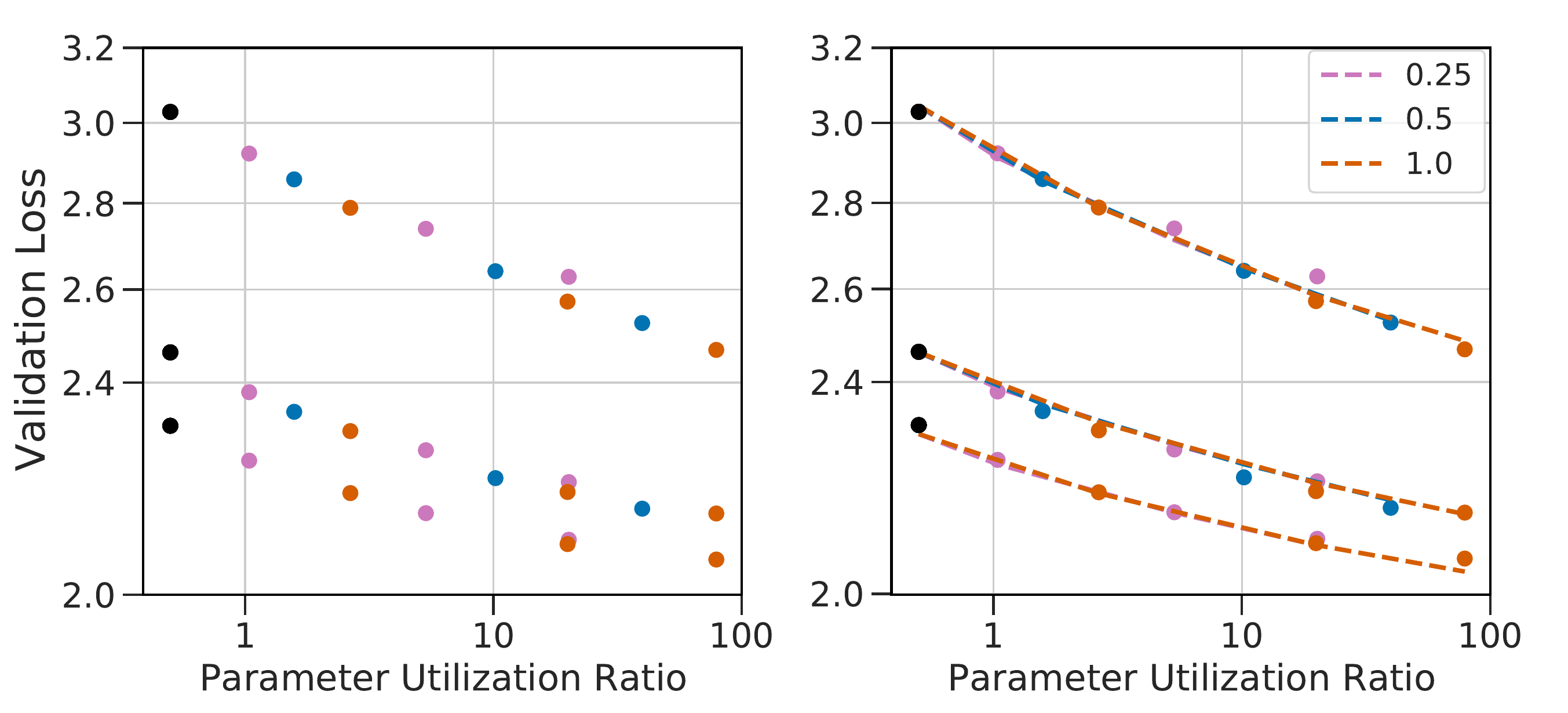}
\caption{Joint fits to Equation \eqref{eq:real_joint_scaling_law_fp} for $R \in \{0.25, 0.5, 1.0\}$.}
\label{apfig:pf_fits_ben_individual}
\end{figure}

\begin{table}[h]
\centering
\footnotesize
\begin{tabular}{llrrrrr}
\toprule
     &    &      a &      b &     c &     d &  RMSE \\
policy & dataset &        &        &       &       &       \\
\midrule
Dense & Validation Set & -0.078 &   &  & 1.063 & 0.014 \\
     & LAMBADA & -0.203 &   &  & 1.952 & 0.039 \\
     & The Pile & -0.102 &   &  & 1.239 & 0.020 \\
     & CC & -0.097 &   &  & 1.133 & 0.041 \\
     & WikiText-103 & -0.090 &   &  & 1.172 & 0.015 \\
     & C4 & -0.066 &   &  & 1.009 & 0.014 \\
Hash & Validation Set & -0.082 & -0.102 & 0.009 & 1.102 & 0.022 \\
     & LAMBADA & -0.213 & -0.167 & 0.015 & 2.049 & 0.051 \\
     & The Pile & -0.111 & -0.161 & 0.014 & 1.325 & 0.023 \\
     & CC & -0.101 & -0.101 & 0.010 & 1.177 & 0.045 \\
     & WikiText-103 & -0.093 & -0.086 & 0.007 & 1.208 & 0.027 \\
     & C4 & -0.070 & -0.088 & 0.008 & 1.045 & 0.021 \\
S-Base & Validation Set & -0.081 & -0.092 & 0.008 & 1.086 & 0.025 \\
     & LAMBADA & -0.211 & -0.152 & 0.012 & 2.020 & 0.048 \\
     & The Pile & -0.110 & -0.117 & 0.008 & 1.309 & 0.028 \\
     & CC & -0.100 & -0.101 & 0.010 & 1.154 & 0.050 \\
     & WikiText-103 & -0.092 & -0.074 & 0.005 & 1.194 & 0.025 \\
     & C4 & -0.068 & -0.081 & 0.007 & 1.031 & 0.024 \\
RL-R & Validation Set & -0.081 & -0.107 & 0.010 & 1.090 & 0.022 \\
     & LAMBADA & -0.212 & -0.190 & 0.016 & 2.030 & 0.051 \\
     & The Pile & -0.110 & -0.149 & 0.012 & 1.320 & 0.030 \\
     & CC & -0.100 & -0.113 & 0.011 & 1.156 & 0.045 \\
     & WikiText-103 & -0.092 & -0.091 & 0.008 & 1.195 & 0.023 \\
     & C4 & -0.069 & -0.092 & 0.009 & 1.033 & 0.022 \\
\bottomrule
\end{tabular}
\label{tab:transfer_coefficients}
\caption{Scaling coefficients for different downstream tasks.}
\end{table}

\end{document}